\documentclass{article}

\usepackage[preprint, nonatbib]{neurips_2021}

\usepackage[utf8]{inputenc} 
\usepackage[T1]{fontenc}    
\usepackage{hyperref}       
\usepackage{url}            
\usepackage{booktabs}       
\usepackage{amsfonts}       
\usepackage{nicefrac}       
\usepackage{microtype}      
\usepackage{xcolor}         
\usepackage{graphicx}
\usepackage{amsmath}
\usepackage{amssymb}
\usepackage{multirow}
\usepackage{wrapfig}
\usepackage{makecell}
\usepackage{pifont}

\usepackage[font=footnotesize]{caption}

\usepackage{multibib}
\newcites{A}{References}

\newcommand{\header}[1]{\vskip -0.03in \noindent{\bf #1}}

\def\mDelta{\boldsymbol{\Delta}}
\def\mD{\mathbf{D}}
\def\mI{\mathbf{I}}
\def\mM{\mathbf{M}}
\def\mU{\mathbf{U}}
\def\mV{\mathbf{V}}
\def\mX{\mathbf{X}}
\def\mY{\mathbf{Y}}
\def\mZ{\mathbf{Z}}
\newcommand{\R}{\mathbb{R}}

\def\br{\boldsymbol{r}}
\def\bo{\boldsymbol{o}}
\def\bh{\boldsymbol{h}}
\def\bm{\boldsymbol{m}}
\def\bx{\boldsymbol{x}}

\newcommand{\cmark}{\ding{51}}
\newcommand{\xmark}{\ding{55}}

\title{Seeing Through Clouds in Satellite Images}
\author{
  Mingmin Zhao\\
  MIT\\
  \texttt{mingmin@mit.edu} \\
  \And
  Peder A. Olsen \\
  Microsoft \\
  \texttt{peolsen@microsoft.com} \\
  \And
  Ranveer Chandra \\
  Microsoft \\
  \texttt{ranveer@microsoft.com } \\
}

\newcommand{\name} {\mbox{SpaceEye}}
\graphicspath{{./figures/}}

\begin{document}
\maketitle

\begin{abstract}
    This paper presents a neural-network-based solution to recover pixels occluded by clouds in satellite images. We leverage radio frequency (RF) signals in the ultra/super-high frequency band that penetrate clouds to help reconstruct the occluded regions in multispectral images. We introduce the first multi-modal multi-temporal cloud removal model. Our model uses publicly available satellite observations and produces daily cloud-free images. Experimental results show that our system significantly outperforms baselines by 8dB in PSNR. We also demonstrate use cases of our system in digital agriculture, flood monitoring, and wildfire detection. We will release the processed dataset to facilitate future research.
\end{abstract}

\section{Introduction}
Observing and monitoring Earth from space is crucial for tackling some of the greatest challenges of the 21st century, such as climate change, natural resources management, and disaster mitigation.
Recent years have witnessed a surge of interest in Earth observation satellites that orbit around the planet and capture information about the Earth's surface. The increased availability of satellite images with high resolution and short revisit time have enabled numerous applications in digital agriculture~\cite{mulla2013twenty, hank2019spaceborne}, environmental monitoring~\cite{manfreda2018use, tucker2000strategies, tralli2005satellite, cunjian2001extracting}, humanitarian assistance~\cite{lang2017earth, voigt2007satellite}, transport and logistics~\cite{kopsiaftis2015vehicle, van2018spacenet}, and many more.


Satellites with optoelectronic sensors passively examine the Earth's surface across visible and infrared bands to capture images. The resulting multispectral image is similar to an RGB image except that it has more channels corresponding to different wavelengths. However, one fundamental challenge in optical satellite images is the occlusion due to clouds.
Approximately 55\% of the Earth's surface is covered in opaque clouds with an additional 20\% being obstructed by cirrus or thin clouds \cite{wylie1989two, wylie2005trends, stubenrauch2013assessment}.
As a result, these cloud pixels are typically detected and discarded before further analysis. 
Cloud occlusions limit the timely detection of changes (e.g., crop emergence, flooding), thereby missing opportunities for early intervention.
The problem is exacerbated by the intermittent and irregular availability of cloud-free pixels, which poses challenges for statistical analysis and machine learning on top of satellite images.


\begin{figure}[t]
\begin{center}
\includegraphics[width=0.8\linewidth]{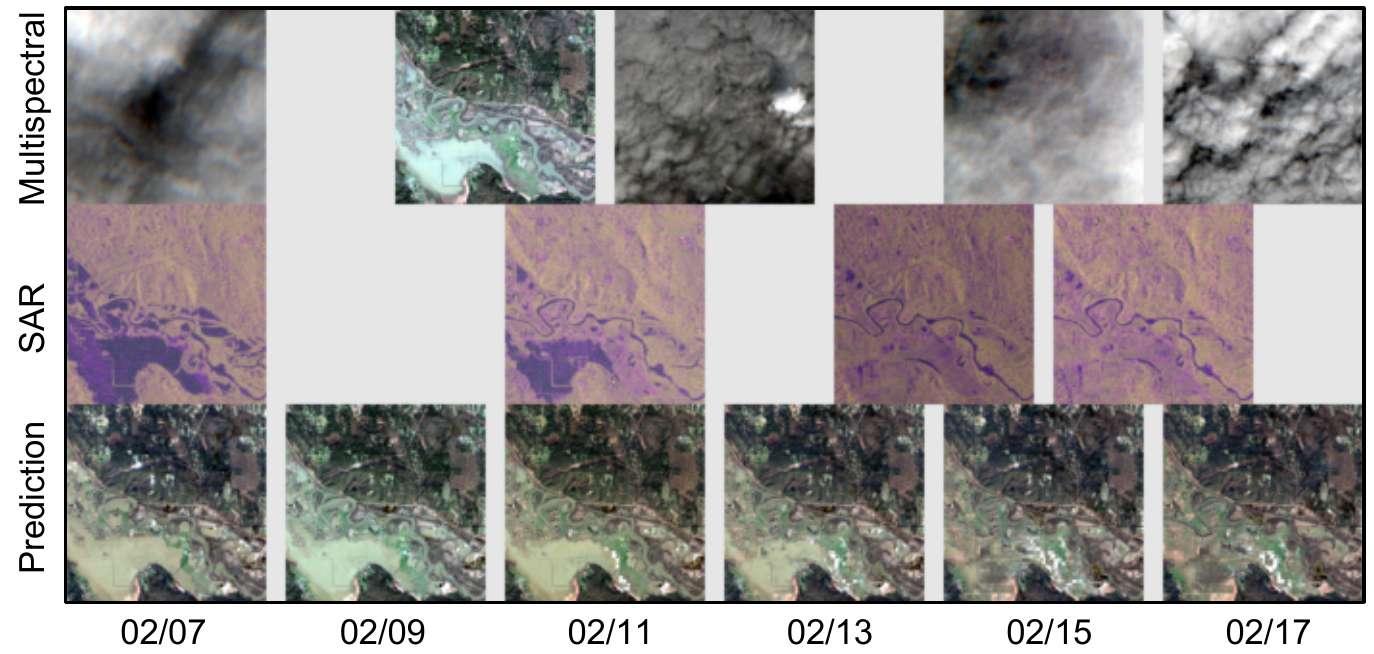}
\end{center}
\vspace{-10pt}
\caption{\footnotesize{
SpaceEye in action: it takes as input multispectral images (top row) and SAR images (middle row) and produces cloud-free predictions (bottom row).  The scene shows a flooded area in Carnation, WA in 2020.  Note the infrequent availability of cloud-free multispectral images during the flooding event. Our model reconstruct the multispectral images covered by clouds, capturing the extent of standing water on February 7th and how the flood drained off after mid-February.
}}\label{fig:teaser}
\vspace{-10pt}
\end{figure}

In this paper, we aim to recover the pixels that are occluded by clouds.
Unlike existing methods on cloud removal~\cite{mahajan2019cloud, enomoto2017filmy, dong2018inpainting, singh2018cloud, lee2019cloud, zheng2020single, tseng2008automatic, xu2016cloud, sarukkai2020cloud}, or more generally, image and video inpainting~\cite{elharrouss2019image, kim2019deep, yu2018generative, yu2019free, liu2018image, iizuka2017globally, pathak2016context, xu2019deep} that generate visually plausible inpainting results based on the context, our goal is to recover optical observations accurately. 
Our main idea is to use a different modality, i.e., radio frequency (RF) signals, to help recover the optical information.
While clouds mostly block visible and infrared wavelengths, RF signals in the ultra/super-high frequencies penetrate clouds. Furthermore, these signals reflect off the Earth's surface. Using a Synthetic Aperture Radar (SAR) technique, one can generate a radar image that captures the surface's radio reflectance even in the presence of clouds or at night.
Existing work has used radar images to detect changes in radio reflectance due to metallic objects~\cite{nunziata2010metallic}, water bodies~\cite{schlaffer2015flood, kussul2011flood}, and geometric transformations of landscape (e.g., constructions, Earthquakes)~\cite{michel1999measuring, knopfle1998mosaicking}.
More recent work has tried to translate SAR image into multispectral images~\cite{wang2019sar, meraner2020cloud}. However, these methods don't consider the temporal history of multispectral images and have limited accuracy~(Table~\ref{tab:stats}).

We introduce \name, a neural-network-based Earth observation system that achieves the best of both optical and RF modalities, i.e., it captures rich information in multispectral bands despite clouds and lighting conditions.
To achieve this, we introduce the first cloud removal model that makes uses of both {\bf multi-modal} and {\bf multi-temporal} satellite observations.
It leverages multi-temporal satellite observations in multispectral and RF bands\footnote{\name~uses publicly available data from the European Space Agency (ESA) governed by the "Legal Notice on the use of Copernicus Data and Service Information."  ESA provides up-to-date multispectral and RF measurements from the different Sentinel missions~\cite{Copernicus, sentinelmission}.} to recover daily cloud-free multispectral images.

Figure~\ref{fig:teaser} shows an example output of our system tracking the floods in a farm located in Carnation, WA.
It shows the sequences of multispectral
and SAR images captured during this period. As can be seen, most of the multispectral images capture clouds instead of the farm underneath them. While SAR images don't capture as rich information as multispectral images could, they provide consistent observation of RF reflectivity on the ground. Our system generates daily cloud-free multispectral predictions as shown in the bottom row. As can be seen, it captures the extent of the standing water on February 7th, despite the fact that the multispectral image on the same day was covered in clouds. It also shows how the flood water gradually drained off after mid-February.


There are several challenges in designing and training our system.
First, effectively fusing multispectral images and RF signals is challenging. These multispectral images and RF signals are captured by different satellites on separate orbits. Yet, the clouds in the multispectral images have a non-predictable and free-form distribution.
Second, there are no ground truth cloud-free images that can be used during training. This poses challenges for supervised training and limits adversarial training, which can help ensure the fidelity of the predictions.

In this paper, we tackle the above challenges as follows. We use a technique called Synthetic Aperture Radar (SAR) to convert RF time signals into 2D images and align them with the multispectral images.
We design a multi-modal attention mechanism to fuse this multi-modal data.
Specifically, we use RF information to decide where each pixel should be attending to when reconstructing the multispectral images.
We also propose a new adversarial training setting that only requires partial observations of examples in the target distribution. This modified adversarial training setting allows us to ensure the fidelity of cloud-free predictions in the absence of ground-truth cloud-free images.





We train on a dataset containing both the multispectral and RF observations spanning 3 years and covering 29 regions of size 100-by-100 km$^2$ across the US and Europe. We test it in various areas from the US, Europe, and Africa.  Experiments show that \name~achieves a peak signal-to-noise ratio (PSNR) of 28.9, which significantly outperforms baselines. We further conduct several case studies to demonstrate the use of \name~to agriculture, wildfire monitoring, and storm damage assessment.









\section{Related Work}


\header{Cloud removal in satellite images.}
Cloud removal starts by detecting cloud pixels~\cite{foga2017cloud, sanchez2020comparison, bai2016cloud, jeppesen2019cloud, zhu2012object, baetens2019validation} and then reconstructs these pixels based on the remaining ones.
The effectiveness of existing cloud removal methods is tightly coupled with the amount of input information.
Existing work generally falls into two categories: single-image and multi-temporal methods.
Single-image methods take a single satellite image as input. 
Various image inpainting algorithms have been used for this task including sparse dictionary learning~\cite{meng2017sparse, li2017removal}, 
and neural-networks-based approaches~\cite{enomoto2017filmy, dong2018inpainting, singh2018cloud, lee2019cloud, zheng2020single}. These algorithms generate visually realistic and semantically correct images in the missing regions. However, inpainting-based methods have limited accuracy for large missing regions, which is common due to the prevalence of clouds.
Multi-temporal methods~\cite{tseng2008automatic, xu2016cloud, sarukkai2020cloud} uses multiple satellite images from nearby days to produce a single cloud-free image, e.g., patch-based methods~\cite{helmer2005cloud, lin2012cloud, vuolo2017smoothing} and tensor factorization approaches~\cite{ji2018nonlocal,aravkin2014variational, ma2017fusion, wang2016removing}.
Several systems~\cite{hazaymeh2015spatiotemporal, luo2018stair, houborg2018cubesat, wang2018spatio} further fuse multispectral images sequences from multiple satellites to increase the number of images in any given time range. The performance of these systems is much improved compared to the single-image methods.
However, these systems assume little change between cloud-free observations and cannot detect abrupt changes.
RF information has also been used as auxiliary information to help remove clouds~\cite{ebel2020multi, meraner2020cloud, wang2019sar}. These methods use the SEN12MS dataset~\cite{schmitt2019sen12ms}, which provides pairs of multispectral and RF images, but do not consider the temporal sequence of satellite images.
Collectively, our work distinguishes itself from previous methods by straddling both the multi-modal and multi-temporal domains, i.e., it utilizes sequences of multispectral and SAR images for cloud removal.
Experimental results show the advantages of our method (Section~\ref{subsec:quantitative}).

\header{Image and video inpainting.}
Both traditional diffusion or patch-based methods~\cite{criminisi2004region, simakov2008summarizing, barnes2009patchmatch, strobel2014flow, huang2016temporally} and learning-based methods~\cite{yu2018generative, yu2019free, liu2018image, iizuka2017globally, pathak2016context, xu2019deep, kim2019deep} have been used for image and video inpainting.   
We build on this literature and adopt a learning-based approach but differ in that we use multi-modal data and aim to recover the actual observations occluded by clouds.
While the attention mechanism has been widely used in the inpainting models~\cite{yu2018generative, yu2019free}, we used auxiliary RF information from the missing region to guide the attention instead of using spatial context from nearby pixels.

\header{Learning with multiple modalities.}
Our work is related to cross-modal and multi-modal learning that leverages complementary information across modalities~\cite{soundofpixels, owens2018audio, gao2018learning, soundofmotions, rfpose3d, SSundaram:2019:STAG, salvador2017learning}.
Multi-modal learning of images and RF have been explored to sense people through walls~\cite{rfpose, li2019making, rfavatar}.
While self-supervised and unsupervised methods were typically used to guide the representation learning, we use multi-modal data to recover missing data and focus on the task of cloud removal.

\section{Synthetic Aperture Radar Imaging}

\begin{wrapfigure}{r}{0.4\textwidth}
  \vspace{-25pt}
  \begin{center}
    \includegraphics[width=0.4\textwidth]{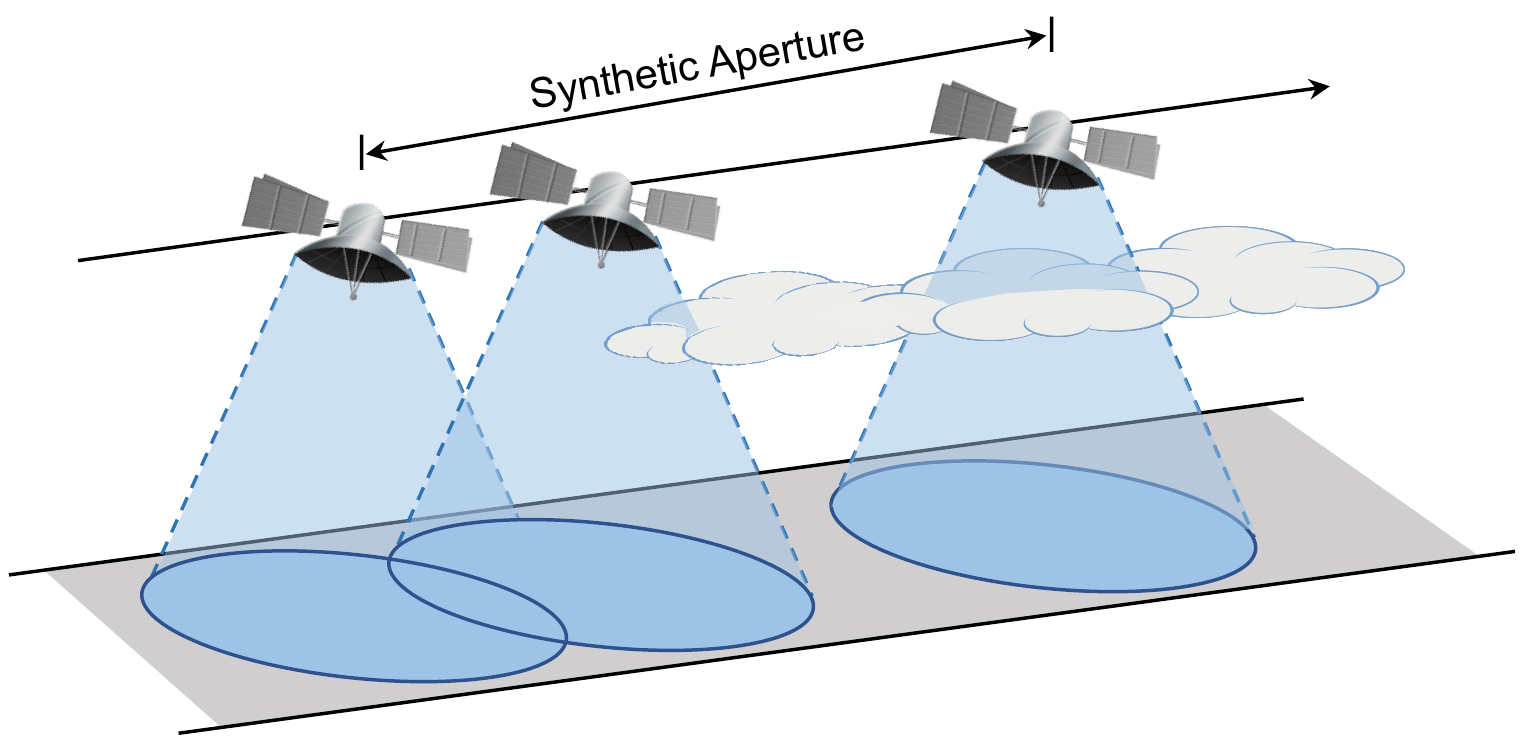}
  \end{center}
  \vspace{-10pt}
  \caption{\footnotesize{
  Basic principle of synthetic aperture radar (SAR) imaging.
  }}\label{fig:sar}
  \vspace{-10pt}
\end{wrapfigure}

Synthetic-aperture radar (SAR) is a remote sensing method that has attracted much interest.
Aircraft or spacecraft equipped with SAR actively transmits pulses of radio waves to ``illuminate'' the target region and record the waves reflected after interacting with the Earth.
The recorded SAR data is very different from optical images, which can be interpreted similarly to a photograph. Radio waves are time signals that are responsive to surface characteristics like structure and moisture.
Radar sensors usually utilize longer wavelengths at the centimeter to meter scale, which penetrates clouds.


Radar imaging algorithms transform the recorded radio data into two-dimensional images. The spatial resolution of the resulting radar image is directly related to the ratio of the antenna size to the signal wavelength. However, to get a spatial resolution of 10 m from a satellite operating at a wavelength of about 5 cm, requires a radar sensor about 4,250 m long, which is not practical.
SAR is a clever workaround that leverages the synthetic aperture. Specifically, SAR uses the satellite's motion over the target region, as shown in Figure \ref{fig:sar}. A sequence of acquisitions from a smaller radar sensor is combined to simulate a much larger antenna, thus providing higher resolution SAR images~\cite{soumekh1999synthetic}.

We use Sentinel-1 SAR products made publicly available by the European Space Agency. We process the raw dual polarisation Level-1 products with the SAR imaging algorithm and other processing steps, including orbit correction, noise removal, calibration, and terrain correction. Please refer to the supplementary materials for a detailed description of the SAR processing.
\section{Method}
\begin{figure}[htbp]
\begin{center}
\includegraphics[width=0.9\linewidth]{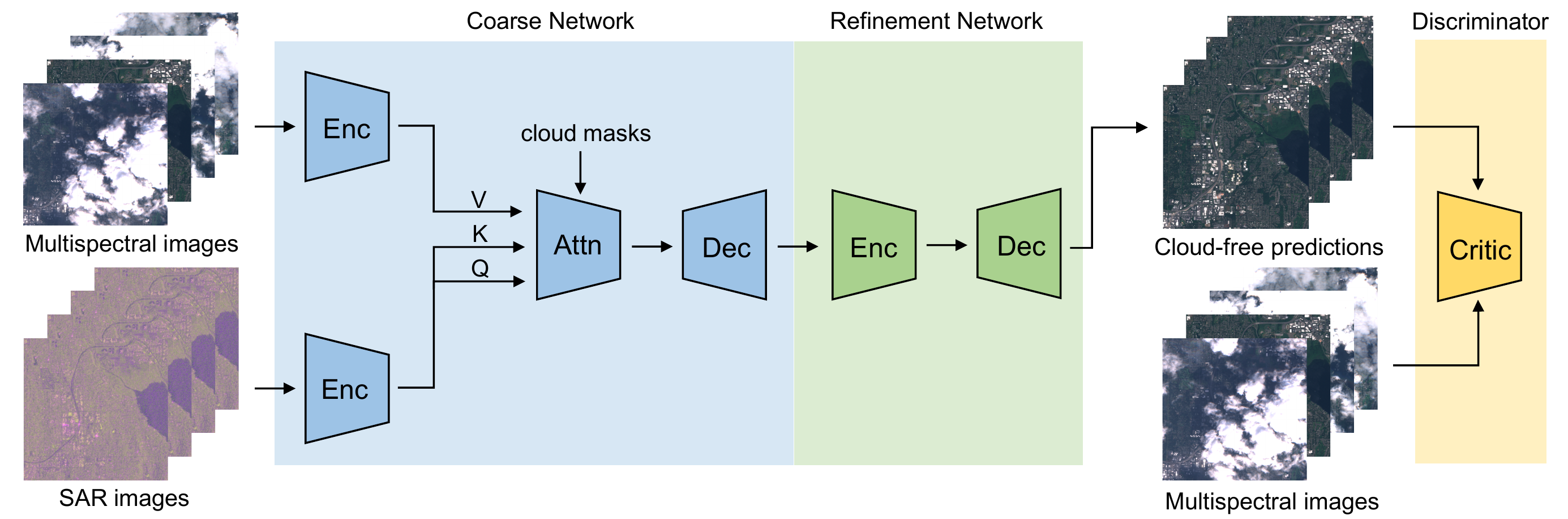}
\end{center}
\vspace{-10pt}
\caption{\footnotesize{
Overview of the model used used in \name.  It inputs multispectral and SAR image sequences to a coarse-to-fine two-stage network with multi-modal attention.
A discriminator with partial observations is used during training to improve the quality of predictions.
}}\label{fig:overview}
\vspace{-10pt}
\end{figure}

We propose a neural network framework that uses multispectral and RF satellite images to reconstruct cloud-free multispectral images.
Figure~\ref{fig:overview} illustrates the model architecture of \name. It takes two streams of satellite images (i.e., multispectral and SAR) as input and uses a coarse-to-fine two-stage network to predict cloud-free images.
In the first stage, a coarse network encodes input multispectral and SAR images and learns to recover the missing pixels via masked attention (Section \ref{subsec:coarse-network}).
In the second stage, a refinement network further improves the predictions through a pair of encoder and decoder with skip connections (Section \ref{subsec:refinement-network}). To ensure the fidelity of the predictions, we propose a new adversarial training setting that trains the discriminator with partial observations, i.e., it is trained with satellite images that are often cloudy (Section \ref{subsec:pogan}).
We describe how we train \name~(Section \ref{subsec:training}) followed by a matrix completion variant of it (Section \ref{subsec:spaceeye-tc})

\subsection{Multi-modal Fusion with Attention}\label{subsec:coarse-network}
The first stage of our model takes the input sequence of multispectral and SAR images and extracts an encoding for each spacetime location. Our model considers the spatial and temporal context to recover the multispectral encoding for spacetime locations that are occluded by clouds.
Specifically, it uses an attention mechanism to effectively borrow information from other possibly distant spacetime locations.
It uses SAR encodings that are available for all spacetime locations to guide the attention mechanism.
This network finally uses a decoder to transform the recovered encoding back to the multispectral space.
We describe each component of the coarse network in detail:

\header{Encoders:} The coarse network uses two encoders for multispectral and SAR images. Each encoder has 9 convolution layers with BatchNorm~\cite{batchnorm} and LeakyReLU~\cite{xu2015empirical}. It encodes each image from the time sequence separately with 3x3 convolution kernels. The encoders reduce the spatial resolution to 1/8 while maintaining the temporal resolution.

\header{Multi-modal attention:}
Our model uses attention mechanism~\cite{attention, vaswani2017attention} to reconstruct the spacetime locations occluded by clouds.
A key component in the attention mechanism is the key and query vectors that allow for the computation of attention scores.
We designed a multi-modal attention mechanism that computes key and query vectors based on the SAR encodings, which are always available despite clouds. 
Moreover, we use masked attention to make sure the model will only learn to attend to cloud-free spacetime locations. The mask we use here is the cloud mask from the \texttt{S2Cloudless} algorithm~\cite{baetens2019validation}. 

Precisely, our neural attention model computes output $\bh$ from SAR encoding $\br$, multispectral encoding $\bo$, and cloud mask $\bm$ (1 for cloud-free pixels and 0 otherwise) as follow:
\begin{equation}
\bh_i = \frac{1}{\mathcal{C}(\br, \bm)} \sum_{\forall j}a(\br_i, \br_j, \bm_j) \cdot V(\bo_j).
\end{equation}
Here $i$ is the index of an output position in spacetime, and $j$ is the index that enumerates all possible positions.
The value vector $V(\cdot)$ is computed using multispectral encodings $\bo$.
The attention scores $a(\cdot)$ is normalized by a factor $\mathcal{C}(\br, \bm)$. We use dot product followed by a Softmax operation to compute the attention scores as:
\begin{equation}
a(\br_i, \br_j, \bm_j) = e^{K(\br_j)^TQ(\br_i)} \cdot \bm_j,
\end{equation}
where the key $K(\cdot)$ and the query $Q(\cdot)$ are computed using SAR encodings $\br$. Note that the dot product score is multiplied by the cloud mask, which forces the attention module to ignore spacetime positions with clouds (i.e., $\bm_j = 0$).

\header{Decoder:} The decoder has 3 up-sampling layers to recover the original spatial resolution. Each up-sampling layer is followed by two 3x3 convolution layers.

\subsection{Refinement Network}\label{subsec:refinement-network}
We follow Yu et al.~\cite{yu2018generative} to design a coarse-to-fine network architecture that progressively improves the quality of the predictions. We use a variant of the U-Net architecture~\cite{ronneberger2015u} with 3D spatial-temporal convolutions.
The {\bf encoder} has 7 down-sampling blocks with 3x4x4 convolution kernels, BatchNorm, and LeakyReLU. The first 3 blocks have a stride of 2x2x2 while the rest use 1x2x2. Each block doubles the number of channels.
The {\bf decoder} has 7 up-sampling blocks with transposed convolution. It also uses skip connection from the intermediate representations of the encoder.

\subsection{Adversarial Training with Partial Observations}\label{subsec:pogan}
We would like to ensure the fidelity of our model's prediction, i.e., we want the predicted cloud-free multispectral images and their temporal evolution to be realistic.
An effective way to ensure fidelity is to use adversarial training~\cite{gan, pix2pix, CycleGAN, yu2018generative}. At the equilibrium state of adversarial training, the generated distribution will converge to the target distribution (i.e., distribution of positive examples).
However, typical adversarial training won't work in our case since the ground truth multispectral images contain clouds that cannot be used during training. 

\begin{wrapfigure}{r}{0.45\textwidth}
  \vspace{-15pt}
  \begin{center}
    \includegraphics[width=0.43\textwidth]{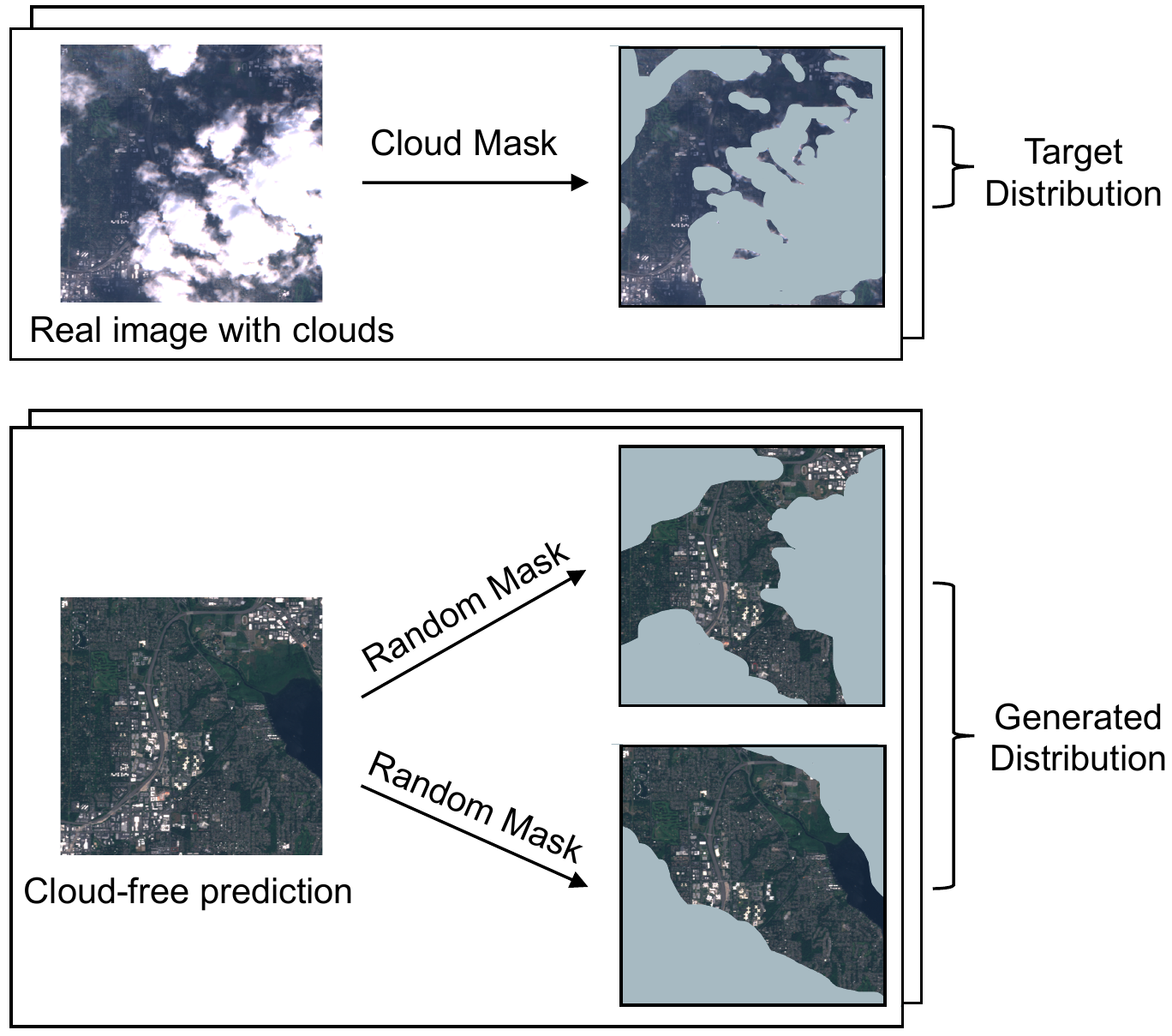}
  \end{center}
  \vspace{-10pt}
  \caption{\footnotesize{
  Training with partial observations.
  }}\label{fig:pogan}
  \vspace{-10pt}
\end{wrapfigure}

To address this issue, we introduce a new adversarial training setting that can learn with partial observations of the target examples.
In this setting, the discriminator no longer has full access to neither positive nor negative examples. Instead, it will only have access to a randomly masked version of the examples.
As illustrated in Figure \ref{fig:pogan}, the discriminator only has access to the cloud-free pixels for positive examples, i.e., they will be masked by the cloud detection results.
On the other hand, the negative examples generated by our decoder will be masked with randomly generated cloud mask before sending to the discriminator.
Mathematically, Equation \ref{eqn:gan} shows the value function for the vanilla adversarial training, and Equation \ref{eqn:pogan} shows the value function for the partial-observation-based adversarial training. $p_{data}$ and $p_{model}$ denote the distribution of real multispectral images and the distribution of the generated cloud-free multispectral images, respectively, $\bm$ denotes the cloud mask, and $D$ is the discriminator.
\begin{equation}
    \mathbb{E}_{\bx\sim{p_{data}}}\text{log}D(\bx) + \mathbb{E}_{\bx\sim{p_{model}}}\text{log}(1 \!-\! D(\bx))
    \label{eqn:gan}
\end{equation}
\vspace{-8pt}
\begin{equation}
    \mathbb{E}_{\bx,\bm\sim{p_{data}}}\text{log}D(\bx\cdot\bm) + \mathbb{E}_{\bx\sim{p_{model}}}\mathbb{E}_{\bm}\text{log}(1 - D(\bx\cdot\bm))
    \label{eqn:pogan}
\end{equation}

Consider the equilibrium states of adversarial training using the objectives defined in Equations~\ref{eqn:gan} and \ref{eqn:pogan}. The vanilla adversarial training reaches equilibrium when $p_{model}$ converges to $p_{data}$. This is, however, not ideal for the task of cloud removal since $p_{data}$ is not cloud-free. On the other hand, the equilibrium state of training with Equation~\ref{eqn:pogan} ensures that randomly sampled patches of the prediction $p_{model}$ converges to the cloud-free patches from $p_{data}$.

We use partial convolution~\cite{liu2018image} as the basic building blocks for our discriminator. It ensures that the output of the convolution operation is properly masked and re-normalized to be conditioned on valid pixels. Our discriminator uses five 3D partial convolutions with 3x4x4 kernels and 1x2x2 strides.

\subsection{Training of \name}\label{subsec:training}
\header{Training with synthetic clouds:} We use synthetic clouds to train the coarse network and refinement network.
Specifically, we mask out (i.e., set to zeros) additional portions of the multispectral images that are cloud-free and use it to provide supervision during training. We generate the synthetic cloud mask by randomly sampling another cloud mask from the training set. We update the cloud mask to the union of the original cloud mask and the newly sampled cloud mask. Note that we don't need to render synthetic clouds in the original image, since all cloudy pixels are set to zeros before being passed to the model.
This training strategy is similar in spirit to the training strategy commonly used in image inpainting models~\cite{liu2018image, yu2018generative, yu2019free}.

\header{Loss functions:} We use $L_1$ loss for the output of the coarse network and the refinement network, comparing them with the held-out synthetic cloud pixels.  We train the discriminator using the loss defined in Equation \ref{eqn:pogan}. The corresponding adversarial loss for the generator is computed by flipping the labels. We add the adversarial loss and the $L_1$ losses with a weight of 20.0 for $L_1$ losses.

\header{Training details:}\label{subsec:trainint}
We randomly crop multispectral and SAR images to the size of 256x256 and use sequences of length 48 spanning 48 days.
We implemented our model in PyTorch. It is trained with the Adam~\cite{kingma2014adam} optimizer for 60,000 iterations. We use a batch size of 4 on two GPUs and an initial learning rate of 2e-5 with 10x decay every 15,000 iterations. The training takes 2 days.

\subsection{A Matrix Completion Variant of \name}\label{subsec:spaceeye-tc}
We provide a matrix completion variant of \name, which we refer to as \name-MC. Intuitively, viewing a time series of satellite images as a big matrix with clouds as missing entries in it, we can recover pixels covered by clouds through matrix completion. We format the matrix such that the first dimension of our matrix is spatial locations, and the second dimension is time$\times$channels (both multispectral and SAR channels). Formatting the matrix this way allows for a low-rank decomposition representing pixels as mixtures of land-types, each of which has its land evolution signature.  


Let $C_1\times T\times H\times W$ and $C_2\times T\times H\times W$ denote the size of the temporal sequence of multispectral and SAR observations.
We concatenate and reshape the tensors above to form the observation matrix $\mY\in\R^{(C_1+C_2) T\times HW}$.
We also use a matrix $\mM$ of the same size to indicate the availability of the entries in the observation matrix $\mY$.
Our goal is to find $\mX$, a low-rank approximation of $\mY$ while maintaining the smoothness of observations over time.
Let $\mX_t\in\R^{(C_1 + C_2)\times HW}$ denote the element-wise time-slices of $\mX$ representing the observations at time t.
We find $\mX$ by solving the following rank-constrained optimization:
\begin{equation}
    \min_{\mathrm{rank}(\mX)\leq \tau}\|\mM \circ (\mX-\mY)\|_F^2 + \alpha \sum_{t=1}^{T-1} \|\mX_{t+1}-\mX_t\|_F^2,
\end{equation}
where $\alpha$ is a damping factor, and $\circ$ is the element-wise multiplication operator. Solving it without the rank constraint decouples channels and becomes \textit{damped interpolation}.  
Both optimization problems can be solved efficiently on a GPU. We outline the details in the supplementary materials.


\section{Dataset}
We extracted and aligned radar and multispectral images using data from ESA's Sentinel-1 and Sentinel-2 missions. The data was downloaded using \href{https://sentinelsat.readthedocs.io/en/stable/}{\texttt{sentinelsat}} licensed under GPLv3+.


\header{Sentinel-1} is a constellation of two satellites that have been operating since April 2016 with a joint location revisit rate of 3 days at the equator and 2 days in Europe.   Sentinel-1 carries a single C-band synthetic aperture radar instrument operating centering 5.405 GHz.  For our experiments, we used data from the interferometric wide (IW) swath mode that covers 250km with dual polarisation VV-VH.  

\header{Sentinel-2} 
is a system of two satellites with a joint revisit time of 5 days that have been in operation since March 2017.  The satellites carry a multispectral instrument (MSI) with 13 spectral channels in the visible, near-infrared (NIR), and short-wave infrared spectral range (SWIR), etc.
We use Level 1 top-of-the-atmosphere reflectance products in our experiments.

\header{Data preparation:}
The data preparation involved handling edge effects, merging images on the same absolute orbit, and marking missing pixels.  The Sentinel-1 radar processing additionally involves thermal noise removal, calibration, terrain correction, etc.  The radar processing was done using \href{https://step.esa.int/main/download/snap-download/}{snap} licensed under GPL3, and the details are described in the supplementary materials.
We scale multispectral images to be in the $[0,1]$ interval and SAR images to be in $[-1,1]$. 



\header{Training and test data:}  For the training data we used 32,000 randomly selected spatio-temporal regions spanning 48 days and images of size  $256\times 256$ from 29 tiles.  The 29 tiles were taken from 
\begin{wrapfigure}{r}{0.25\textwidth}
  \vspace{-10pt}
  \begin{center}
    \includegraphics[width=0.2\textwidth]{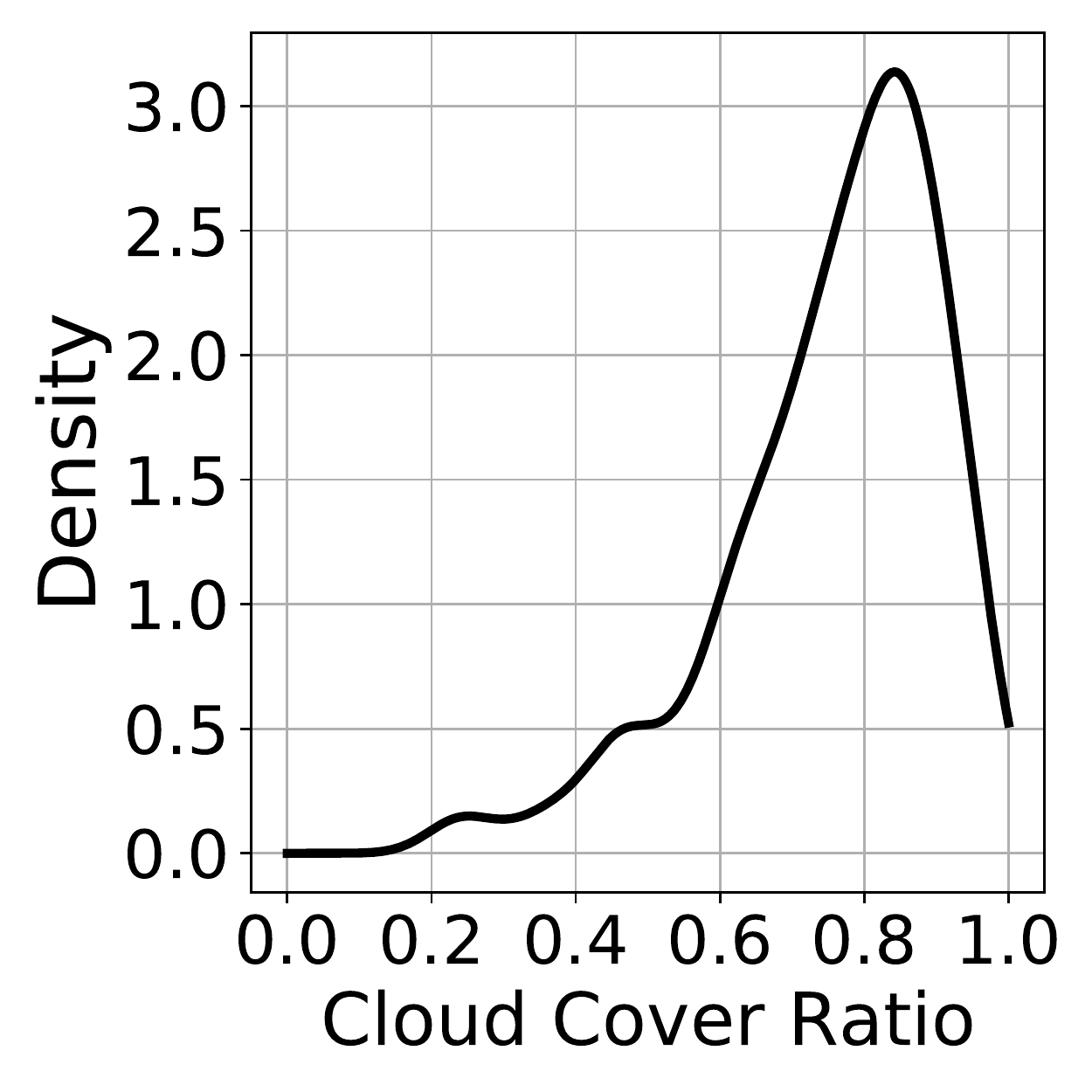}
  \end{center}
  \vspace{-15pt}
  \caption{\footnotesize{Cloud ratio distribution in our dataset.}}
  \vspace{-15pt}\label{fig:cloud_pdf}
\end{wrapfigure}
Europe, Iowa, and Washington between January 2018 until August 2020.  For the validation test we used 30 spatio-temporal regions of the same size from 3 other tiles and for testing we used 1000 randomly selected spatio-temporal regions spanning 48 days and images of size $448\times 448$ from non-overlapping tiles taken from Washington, Iowa, Europe, Rwanda and Australia.
Figure~\ref{fig:cloud_pdf} shows the probability distribution of the cloud cover ratio for the test data, i.e., 48-day sequences of multispectral images. 
There is not a single sequence that is cloud-free, and in fact, a very low fraction of the data has a cloud cover below 50\%. 

\section{Evaluation}
In this section we show qualitative and quantitative results of \name, compare it with several baseline algorithms, and demonstrate example applications of it.

\subsection{Qualitative Evaluation of \name}
\begin{figure*}[h]
\centering
  \includegraphics[width=0.98\linewidth]{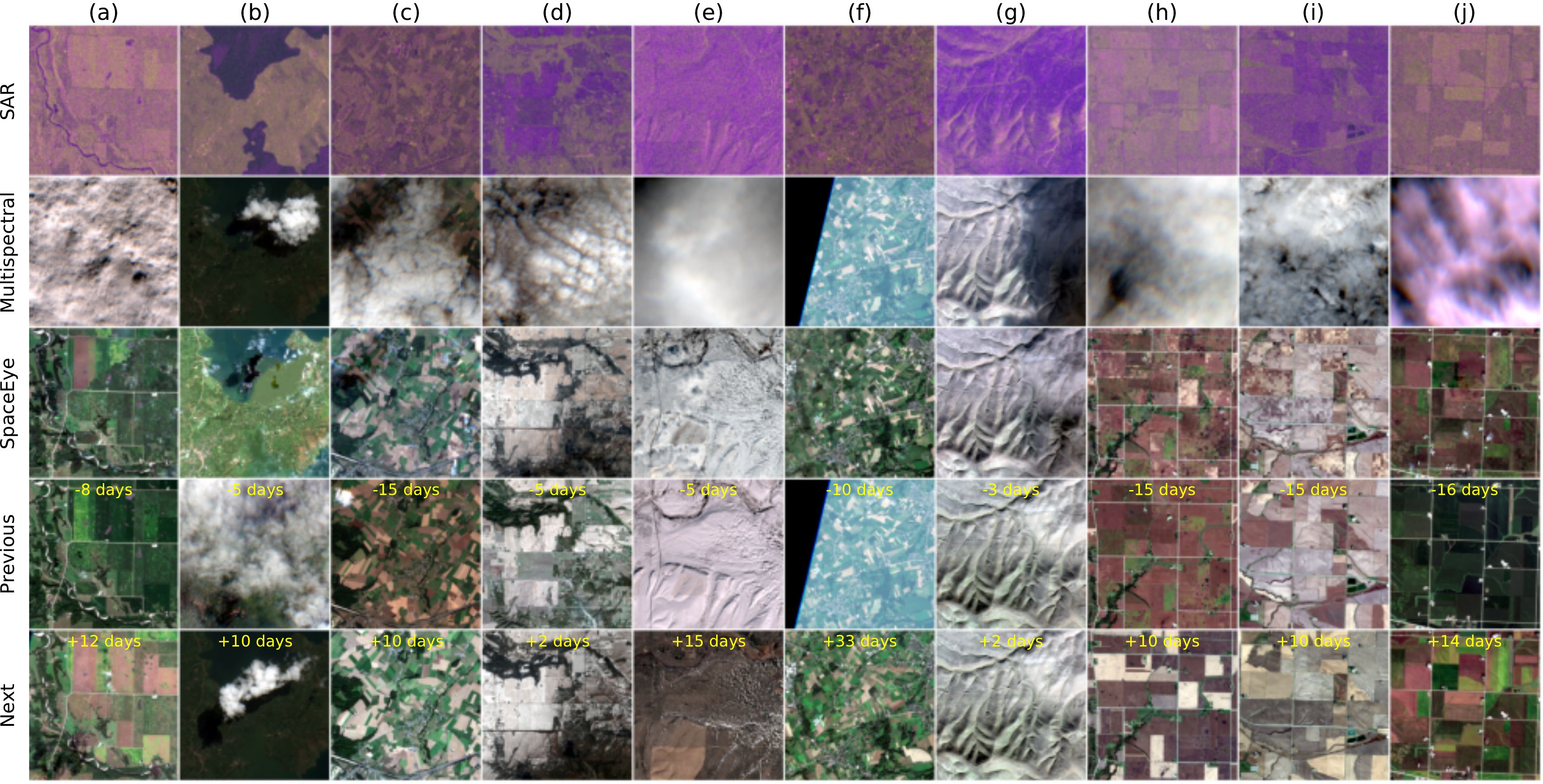}
\caption{\footnotesize{
\name~results on the test set. The first 3 rows show the SAR image, multispectral image, and \name's prediction of the same day.
The last two rows labeled show the nearest cloud-free or the least cloudy multispectral image before and after. The labels on the images show the time relative to the inference time.
}}
\label{fig:collage}
\vspace{-10pt}
\end{figure*}

Figure~\ref{fig:collage} shows the results of \name~on the test data. The first 3 rows show the SAR image, multispectral image, and our prediction of the same day. Note that time sequences of multispectral and SAR images are used as input. To get a sense of what the rest of the images might look like and help understand the prediction, we visualize the nearest cloud-free or the least cloudy multispectral image before and after inference time. Comparing the last 3 rows, our model generates accurate and semantically correct cloud-free images for various land types, including farmlands, hills, villages, etc.
Moreover, our model accurately tracks the changes on the ground. For example, Figure~\ref{fig:collage}(a) shows how it captures harvesting activities: while no fields were harvested in the previously available image and most of them are harvested in the next available image, our prediction shows the field that was first harvested. Similarly, our model captures the changes due to growing plants and agriculture products in Figure~\ref{fig:collage}(h-j), as well as dissolving snow in Figure~\ref{fig:collage}(e)(g). Figure~\ref{fig:collage}(f) shows another example of how our model deals with the cirrus cloud.

\subsection{Quantitative Evaluation of \name}\label{subsec:quantitative}


\header{Baselines.}
We consider two multi-temporal baselines, namely \name-MC and damped interpolation (Section \ref{subsec:spaceeye-tc}). We use $\tau=35$ and $\alpha=3$ for \name-MC and $\alpha=0.5$ for damped interpolation.
We also consider 3 single-image-based baselines: a Cycle-GAN based method that translates a SAR image into a multispectral image~\cite{wang2019sar} (SAR2optical), an image inpainting model~\cite{yu2019free} retrained with multispectral images (DeepFill v2), and a SAR-optical fusion~\cite{meraner2020cloud} method that uses a SAR image together with a multispectral image to remove clouds (DSen2-CR).

\header{Evaluation metrics.} 
To evaluate the cloud removal performance, we used the concept of synthetic clouds where we hold out (i.e., set to zeros) parts of the clear images. We report reconstruction error on synthetic clouds (\texttt{syn}), as well as reconstruction error on synthetic clouds and cloud-free pixels (\texttt{all}).  
The metrics we report are: peak signal-to-noise-ratio (PSNR), mean absolute error (MAE), and Pearson correlation coefficient ($R^2$). We evaluated the metrics on the entire test set.



\header{Results.}
Table~\ref{tab:stats} shows the performance of \name~and baselines across all 10 spectral bands.
As shown in the tables, \name~significantly outperforms all the baselines.
We can also observe that all 3 multi-temporal methods perform better than the 3 methods that do not use temporal history. Among these single-image methods, DSen2-CR~\cite{meraner2020cloud}, which uses both optical and RF inputs, has the best performance.
Table~\ref{tab:bands} breaks down the PSNR for each individual spectral band. The table shows that the PSNR is lower for the red, vegetation red-edge bands and near-infrared bands, as these bands have a higher complexity affected by many plant traits (e.g., leaf chlorophyll and nitrogen content). 

\begin{table}[ht]
\footnotesize
    \centering
\setlength\tabcolsep{3.0pt}
    \begin{tabular}{lcccccccc}
    \toprule
\multirow{2}[3]{*}{Methods} & \multirow{2}[3]{*}{\makecell{input\\modalities}} & \multirow{2}[3]{*}{\makecell{multi-\\temporal?}} & \multicolumn{3}{c}{\texttt{all}} & \multicolumn{3}{c}{\texttt{syn}}\\
\cmidrule(lr){4-6} \cmidrule(lr){7-9}
& & & PSNR & MAE & $R^2$ & PSNR & MAE & $R^2$ \\
\midrule
     \name~(Ours)                     & optical \& RF & \cmark & \textbf{34.84} & \textbf{0.007} & \textbf{0.960} & \textbf{29.92} & \textbf{0.015} & \textbf{0.866} \\
     \name-MC                         & optical \& RF & \cmark & 26.23 & 0.019 & 0.926 & 21.31 & 0.040 & 0.759 \\
     Damped Interp.            & optical       & \cmark & 26.93 & 0.014 & 0.938 & 21.20 & 0.041 & 0.752 \\
     DSen2-CR~\cite{meraner2020cloud} & optical \& RF & \xmark & 26.05 & 0.020 & 0.924 & 20.45 & 0.053 & 0.681 \\
     DeepFill v2~\cite{yu2019free}    & optical       & \xmark & 24.07 & 0.019 & 0.887 & 18.20 & 0.074 & 0.589 \\
     SAR2optical~\cite{wang2019sar}   & RF       & \xmark & 19.43 & 0.056 & 0.593 & 19.91 & 0.054 & 0.621 \\
    \bottomrule
    \end{tabular}
    \vspace{3pt}
    \caption{\footnotesize{Evaluation metrics on the test set.}}
    \label{tab:stats}
    \vspace{-10pt}
\end{table}

\begin{table*}[ht]
\footnotesize
\centering
\setlength\tabcolsep{3.0pt}
\begin{tabular}{lcccccccccc}
\toprule
\multirow{2}{*}{Methods} & 2 & 3 & 4 & 5 & 6 & 7 & 8 &  8A & 11 & 12 \\
& Blue & Green & Red & VRE & VRE & VRE & NIR & NNIR & SWIR & SWIR \\ 
\midrule
\name~(Ours)                    & \textbf{35.63} & \textbf{35.81} & \textbf{34.65} & \textbf{34.89} & \textbf{34.10} & \textbf{33.41} & \textbf{33.37} & \textbf{33.12} & \textbf{37.51} & \textbf{39.72} \\
\name-MC                     & 26.53 & 27.09 & 26.32 & 26.24 & 25.39 & 24.89 & 24.98 & 24.73 & 29.01 & 29.99 \\
Damped Interp.                  & 27.15 & 27.63 & 26.90 & 26.88 & 25.99 & 25.41 & 25.65 & 25.23 & 30.38 & 33.08 \\
DSen2-CR~\cite{meraner2020cloud} & 26.70 & 27.09 & 25.89 & 26.33 & 25.65 & 24.45 & 24.51 & 24.20 & 28.55 & 30.89 \\
DeepFill v2~\cite{yu2019free}   & 25.34 & 25.52 & 23.90 & 24.26 & 24.25 & 23.35 & 23.47 & 22.95 & 23.08 & 25.61 \\
SAR2optical~\cite{wang2019sar}  & 19.20 & 19.68 & 18.77 & 18.99 & 18.78 & 18.36 & 18.50 & 18.41 & 22.46 & 24.76 \\
\bottomrule
\end{tabular}
\vspace{-3pt}
\caption{\footnotesize{PSNR on individual spectral band for different methods on \texttt{all} synthetic clouds and cloud-free pixels (VRE=vegetation red edge, NIR=near infrared, NNIR=narrow band NIR, SWIR=short wave infrared).}}
\label{tab:bands}
\vspace{-5pt}
\end{table*}

\begin{figure}[ht]
\centering
  \includegraphics[width=1.\linewidth]{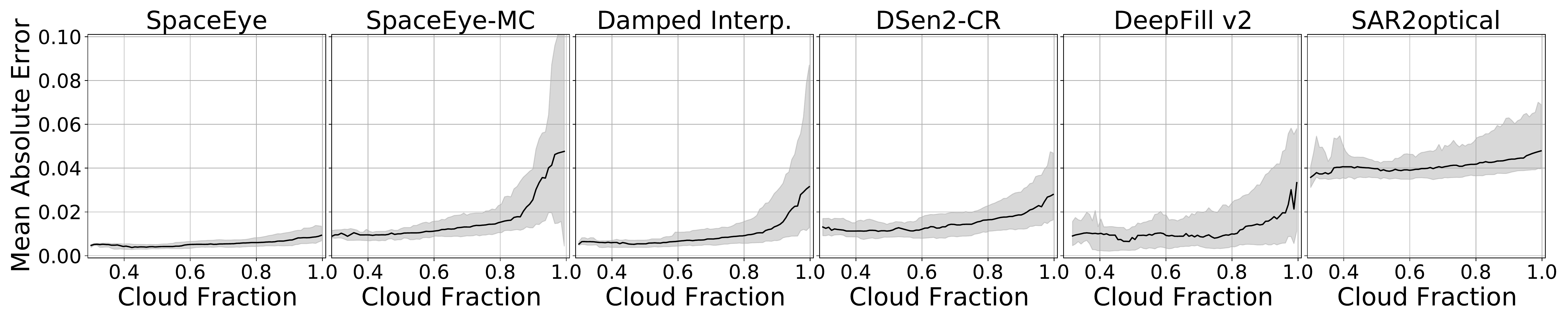}
\caption{\footnotesize{Mean Absolute Error (MAE) for \texttt{all} cloud-free pixels as a function of percentage of cloudy pixels.  The shaded area shows the 25-75\% quantile and the solid line the median for each method.}}
\label{fig:cloud_quantiles}
\vspace{-10pt}
\end{figure}

\header{Performance vs. cloud cover ratio.}
We look at the performance of each algorithm as a function of the cloud cover ratio.  We grouped 1000 test entries according to the cloud cover ratio and computed their MAE.
Figure~\ref{fig:cloud_quantiles} plots the median for each method for cloud-fractions ranging from 0.3 to 0.95.
The shaded area additionally shows the 25\%-75\% quantiles of the error. 
For multi-temporal methods, the median MAE and its variation increase with the cloud cover ratio.
The single scene methods using RF, on the other hand, are less affected by the increased cloud cover ratio. Interestingly, DSen2-CR outperforms both \name-MC and damped interpolation when the cloud ratio is very large, showing the advantages of neural network training over optimization-based methods.
Among all the methods, \name~is robust in the presence of clouds with a smaller shaded area and provides the best results uniformly.  

\subsection{Case Studies of \name~Applications}
We conduct case studies to demonstrate the applications of SpaceEye.

\begin{wrapfigure}{r}{0.55\textwidth}
  \vspace{-20pt}
  \begin{center}
    \includegraphics[width=0.52\textwidth]{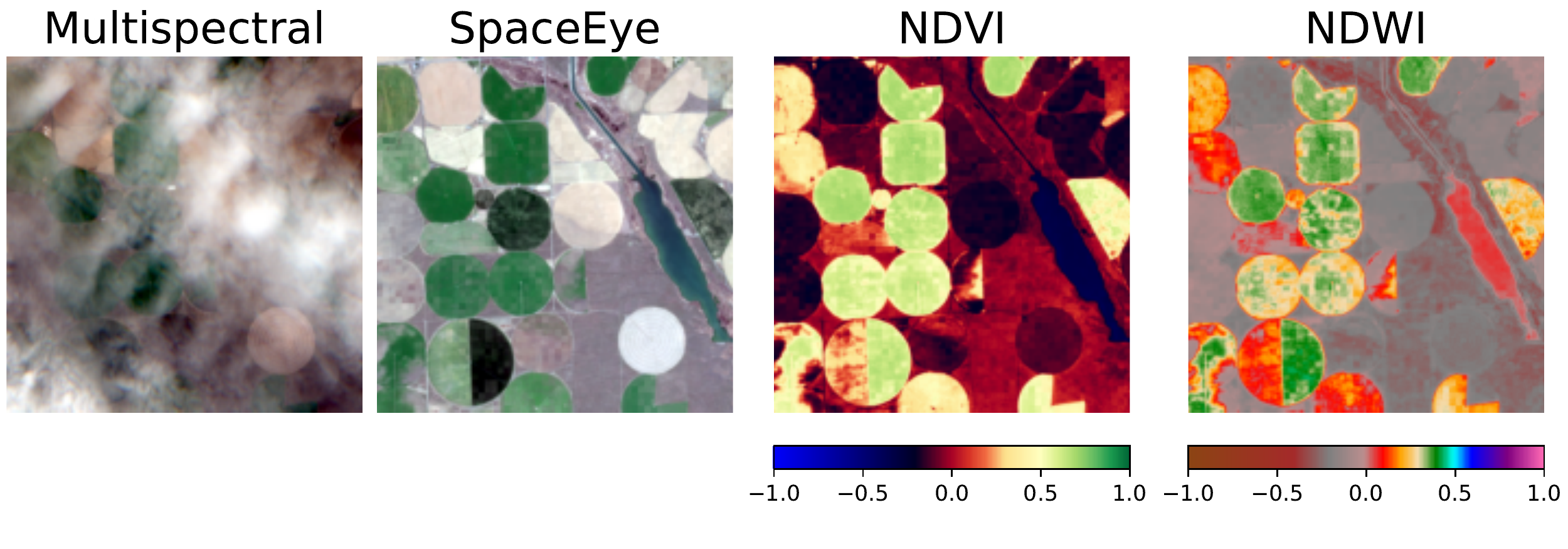}
  \end{center}
  \vspace{-18pt}
  \caption{\footnotesize{\name~inference in Quincy, WA.  The plots shows the RGB prediction along with the NDVI and NDWI.}}
  \vspace{-10pt}\label{fig:ndvi}
\end{wrapfigure}

\header{Case Study I: Digital Agriculture.}
In Figure~\ref{fig:ndvi} we show an application of satellite imaging to agriculture.  We calculate the normalized difference vegetation index (NDVI) used as a qualitative indication of crop health and vitality and the normalized difference water index (NDWI) used to monitor leaf water content.  To calculate NDVI and NDWI, the NIR and SWIR bands are used.  These spectral bands are also produced by \name. In the images we can see in-field variations of these indices.  Normally, NDVI and NDWI would only be calculated on a cloud-free satellite image, but \name~ can provide them even on cloudy days, thereby enabling timely interventions for growers, such as for irrigation, pesticides, nutrients, and other farm management practices.

\begin{wrapfigure}{r}{0.55\textwidth}
  \vspace{-20pt}
  \begin{center}
    \includegraphics[width=0.52\textwidth]{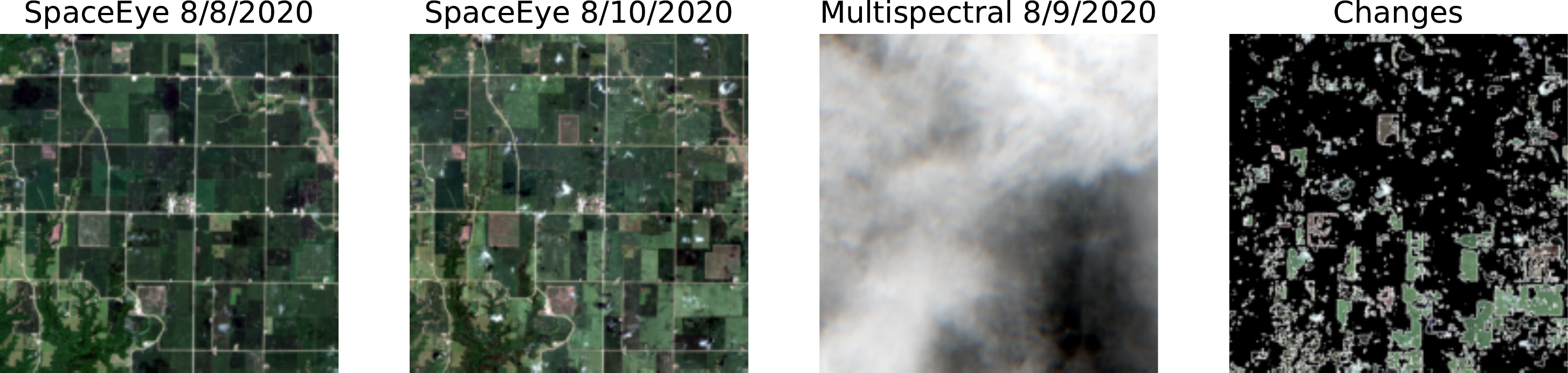}
  \end{center}
  \vspace{-10pt}
  \caption{\footnotesize{Change detection applied to an image from Luther, Iowa before and after a derecho storm on 8/9/2020.  
  }}
  \vspace{-10pt}\label{fig:derecho}
\end{wrapfigure}

\header{Case Study II: Storm Damage Assessment.}
On 8/9/2020, a derecho swept through the state of Iowa and caused billions of dollars of damage.
In Figure~\ref{fig:derecho} we apply \name~ to detect changes due to the storm.  By using \name~ we do not need to wait for a completely cloud-free multispectral image to do the damage assessment. This can trigger a timely response from insurers and the government.

\begin{wrapfigure}{r}{0.38\textwidth}
  \vspace{-20pt}
  \begin{center}
    \includegraphics[width=0.25\textwidth]{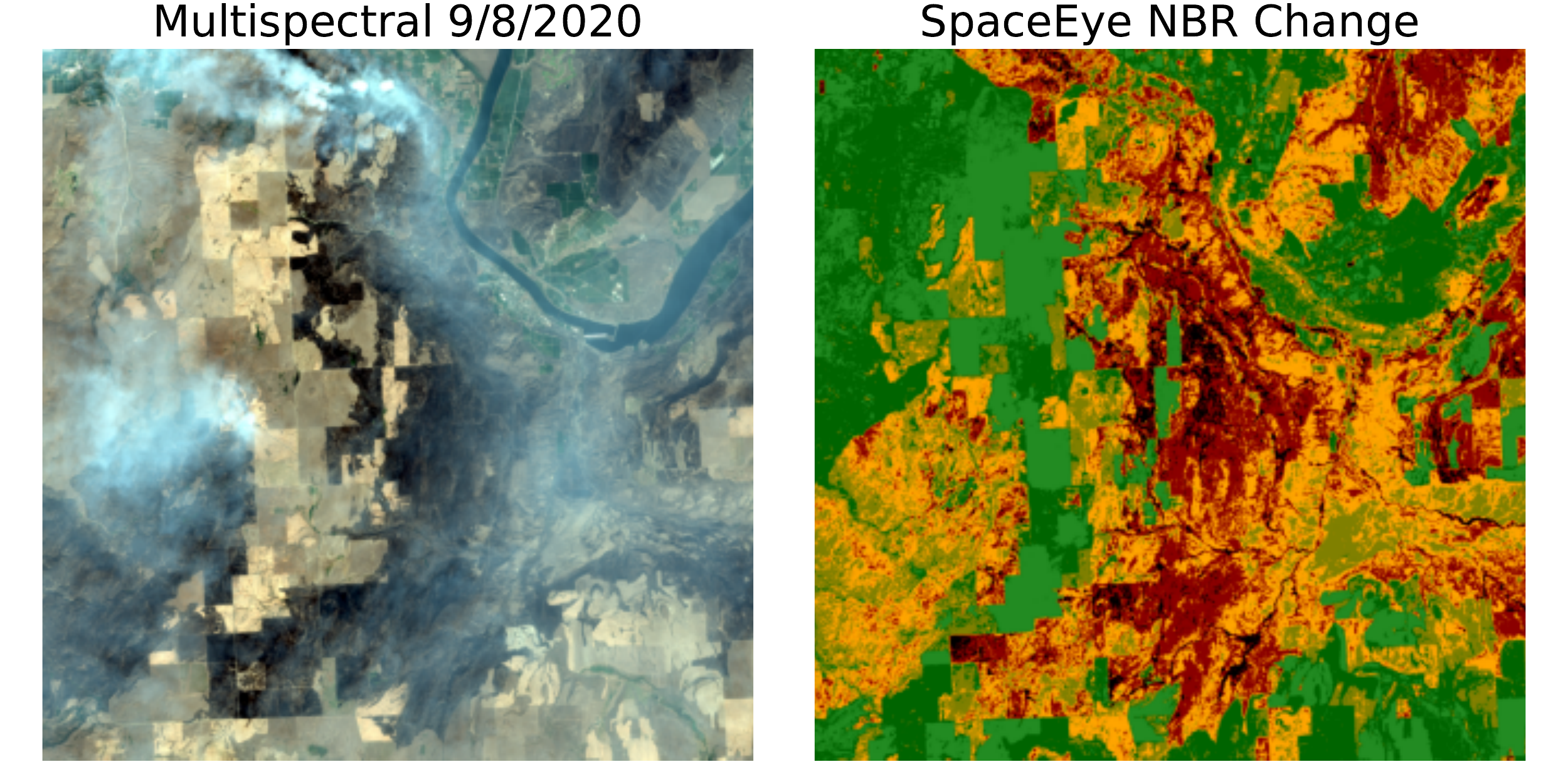}
  \end{center}
  \vspace{-10pt}
  \caption{\footnotesize{Wildfire detection with Normalized Burn Ratio (NBR).}}
  \vspace{-10pt}\label{fig:fire}
\end{wrapfigure}

\header{Case Study III: Wildfire Monitoring.}
Figure~\ref{fig:fire} shows a multispectral image of the wildfire in Bridgeport, WA from September 2020 along with the change in the normalized burn index comparing with the previous year using \name.  The change in the normalized burn index can be used to monitor the burned area and the severity of the burn.

\subsection{Limitations}
The system is limited in that it cannot handle extremely cloudy situations nor gracefully recover from cloud-detection errors.  When there is a lack of cloud-free patches the attention mechanism cannot accurately reconstruct the cloudy data.
This can, however, be addressed by increasing the attention window in time or space dynamically according to the amount of clouds present.  Due to the attention mechanism, failures in the cloud-detector can propagate beyond its original location.
This is an effect we have observed for large water bodies where the cloud-detector is less accurate.


\section{Summary \& Future Work}
This paper presents a new technology to see through clouds in satellite imagery. 
It recovers daily cloud-free multispectral images with high accuracy by learning from multi-modal and multi-temporal satellite observations, leading to 8dB improvement compared to several baseline algorithms.
SpaceEye builds on a growing body of work in multi-modal sensing \& learning. It shows that RF signals, a sensing modality with intrinsically different properties than visible light, can augment vision systems with powerful capabilities. 
This also marks an important step in Earth observation and its applications. We use computer vision to solve a fundamental challenge in remote sensing, which significantly improves the quality of satellite images. SpaceEye also enables a suite of new applications, including timely interventions for agriculture, disaster assessment, and wildfire response. 

We believe \name~opens up exciting opportunities for new applications that could be built on top of these daily cloud-free satellite images.
Moving forward, we are partnering with the government, insurance, and agriculture companies to develop new applications of \name, including forest health monitoring, damage assessment from extreme weather events, wildfires prediction, precision agriculture, and measurement of greenhouse gas from farms.


\bibliographystyle{abbrv}
\bibliography{references}

\newpage
\clearpage
\appendix

This supplementary material is organized as follows:  In Section~\ref{seq:quality} we show \name~predictions for a farm and a farming area in Washington state. We also visualize the normalized difference vegetation index (NDVI), and the normalized difference water index (NDWI) derived from \name~predictions for the entire growing season.  In Section~\ref{sec:data} we describe the detailed preprocessing of Sentinel-1 and Sentinel-2 data needed for efficient training and inference with \name.  We also discuss the details and related work of the cloud-and-cloud-shadow detector used in the system.  Finally, Section~\ref{subsec:tc} discusses the damped interpolation and \name-MC methods and derives the formulas needed to minimize the objective function.

\section{Qualitative Comparison}\label{seq:quality}
\begin{figure*}
\centering
  \includegraphics[width=\linewidth]{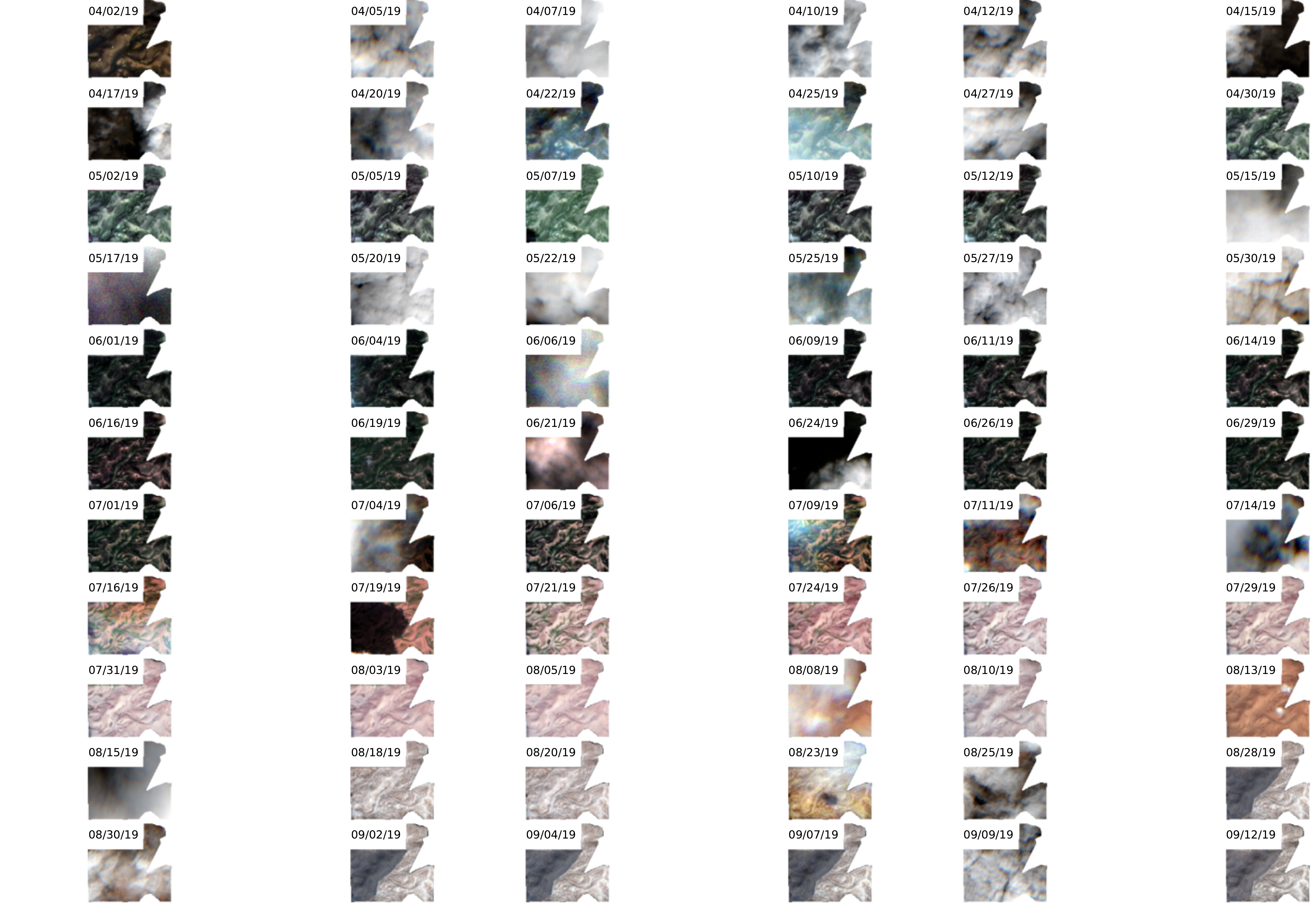}
\caption{
Sentinel-2 images for a hilly farm for the 2019 growing season near Pullman, Washington.  Only pixels belonging to the field is shown.
}
\label{fig:butte_s2}
\end{figure*}

\begin{figure*}
\centering
  \includegraphics[width=\linewidth]{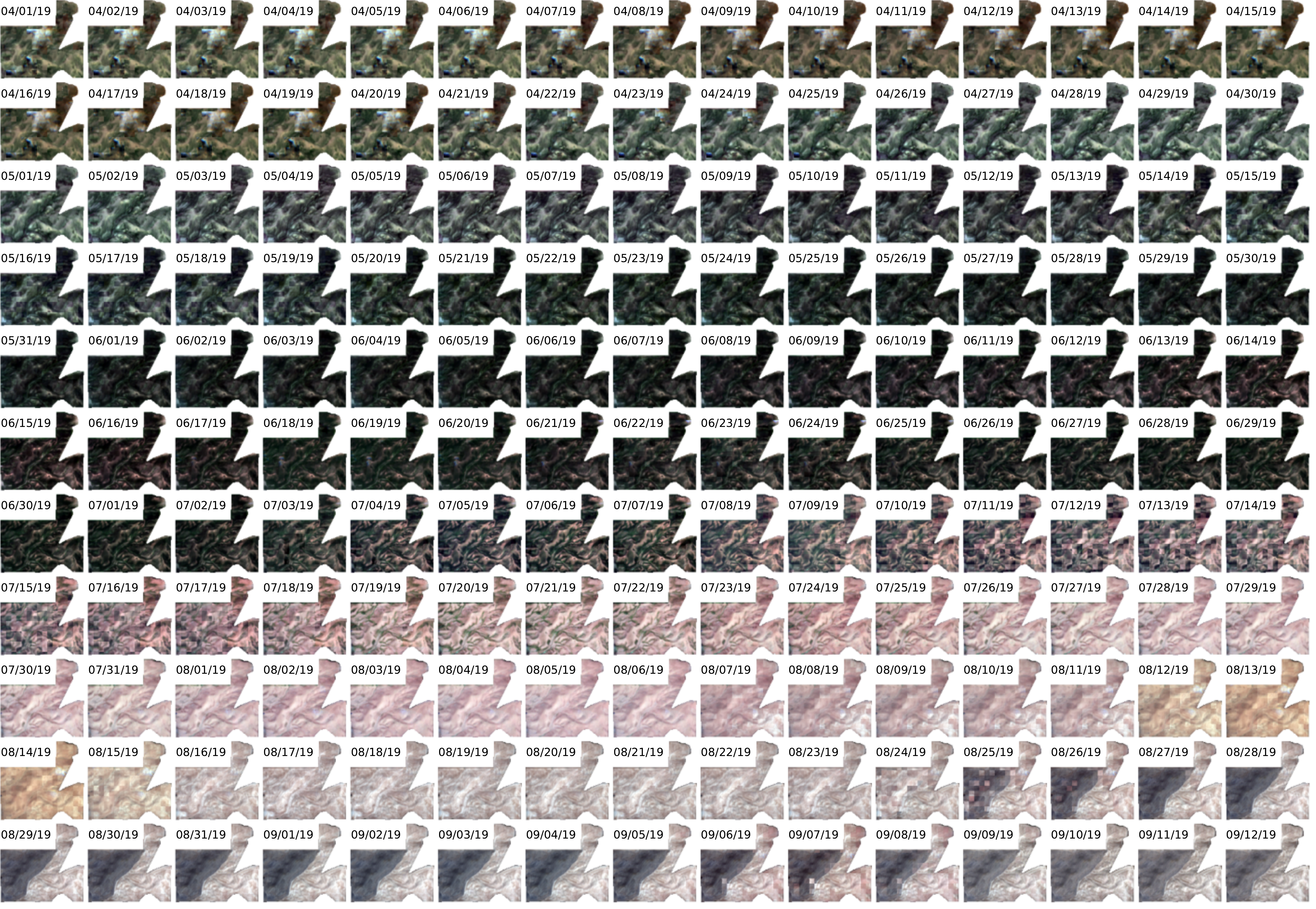}
\caption{
VIR predictions with \name~ for a hilly farm for the 2019 growing season near Pullman, Washington.  Only pixels belonging to the field is shown.
}
\label{fig:butte_fake}
\end{figure*}

\begin{figure*}
\centering
  \includegraphics[width=\linewidth]{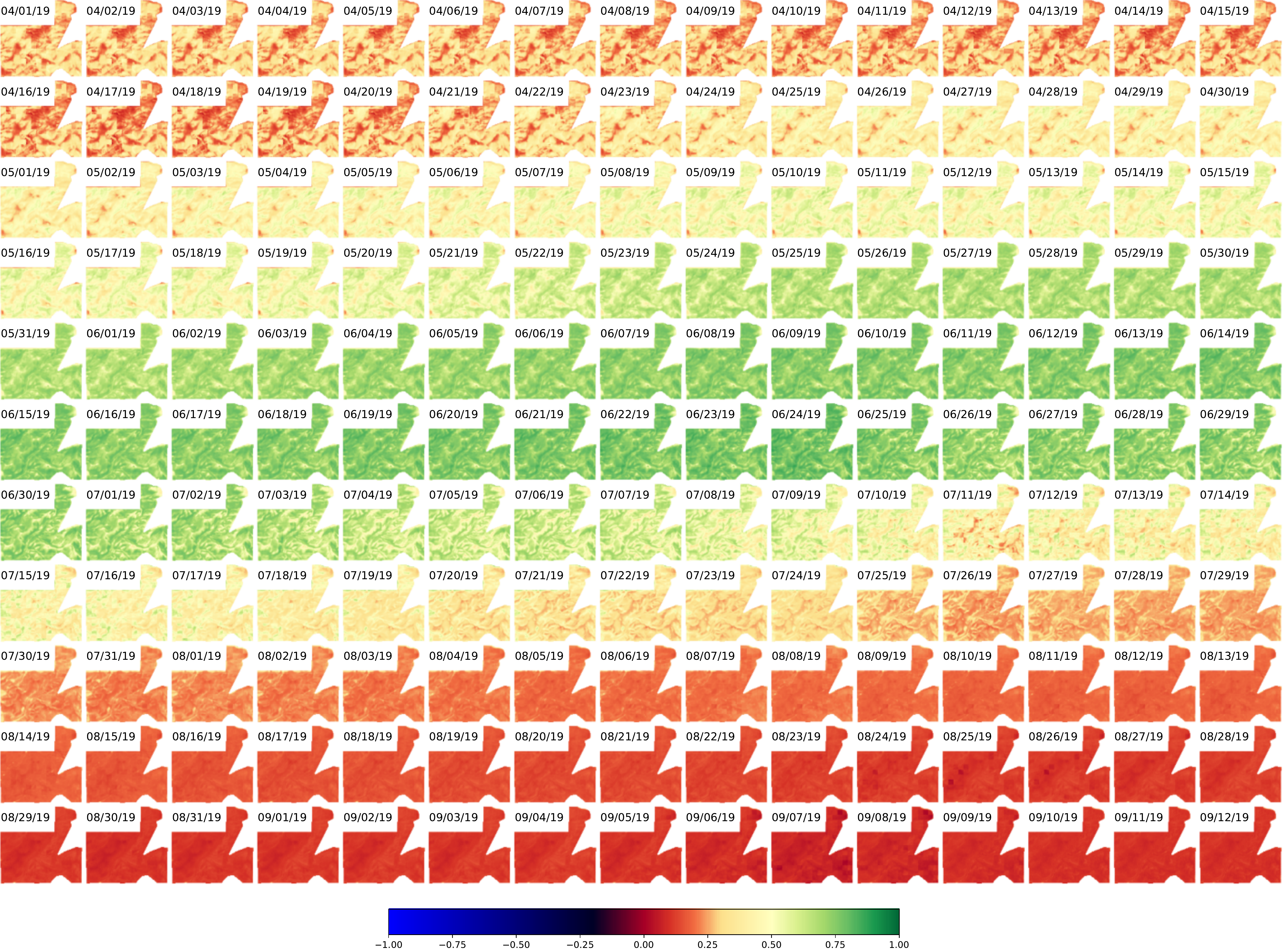}
\caption{
NDVI predictions with \name~ for a hilly farm for the 2019 growing season near Pullman, Washington.  Only pixels belonging to the field is shown.
}
\label{fig:butte_ndvi}
\end{figure*}

\begin{figure*}
\centering
  \includegraphics[width=\linewidth]{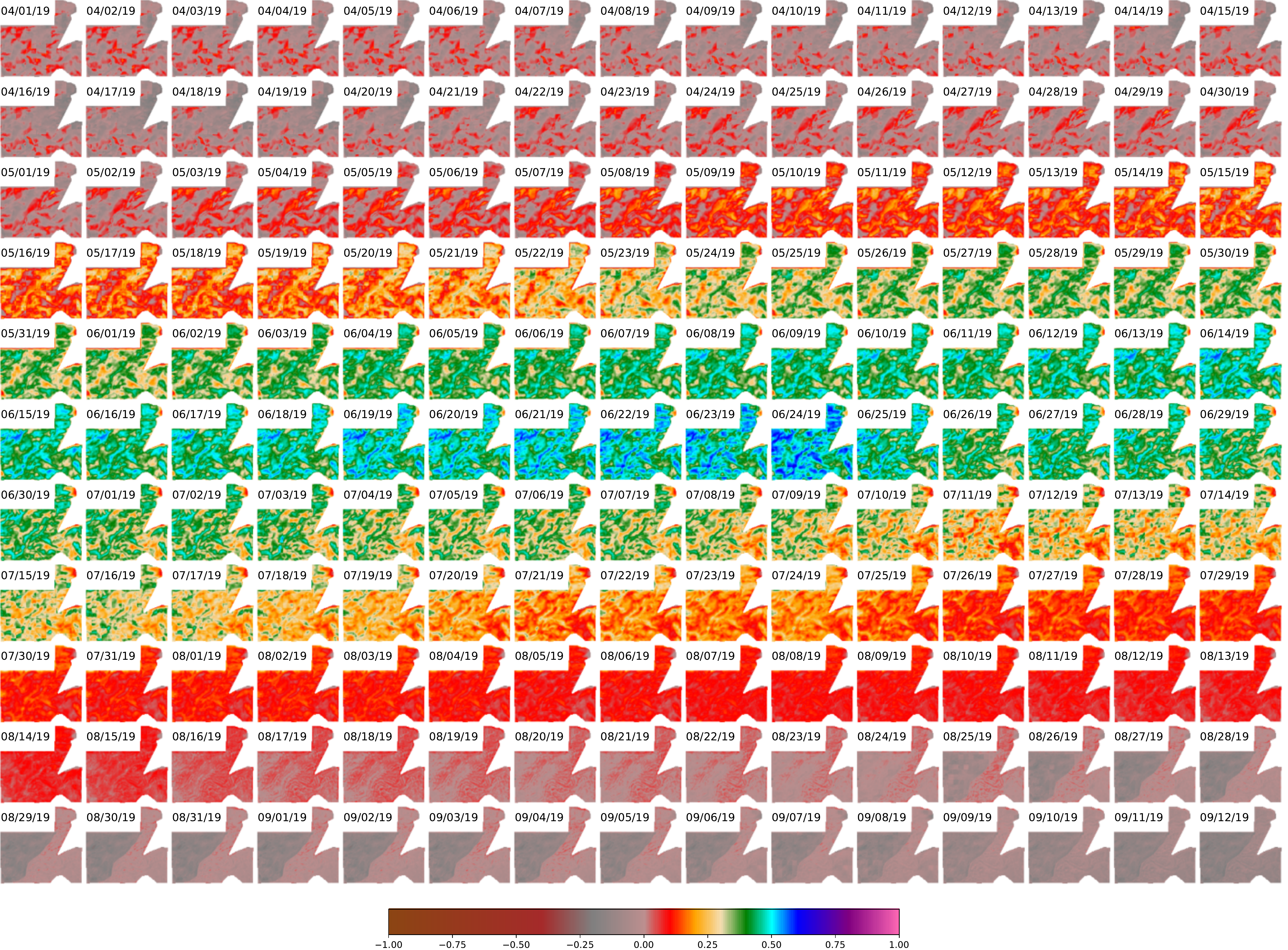}
\caption{
NDWI predictions with \name~ for a hilly farm for the 2019 growing season near Pullman, Washington.  Only pixels belonging to the field is shown.
}
\label{fig:butte_ndwi}
\end{figure*}

\begin{figure}
\centering
\includegraphics[width=0.4\linewidth]{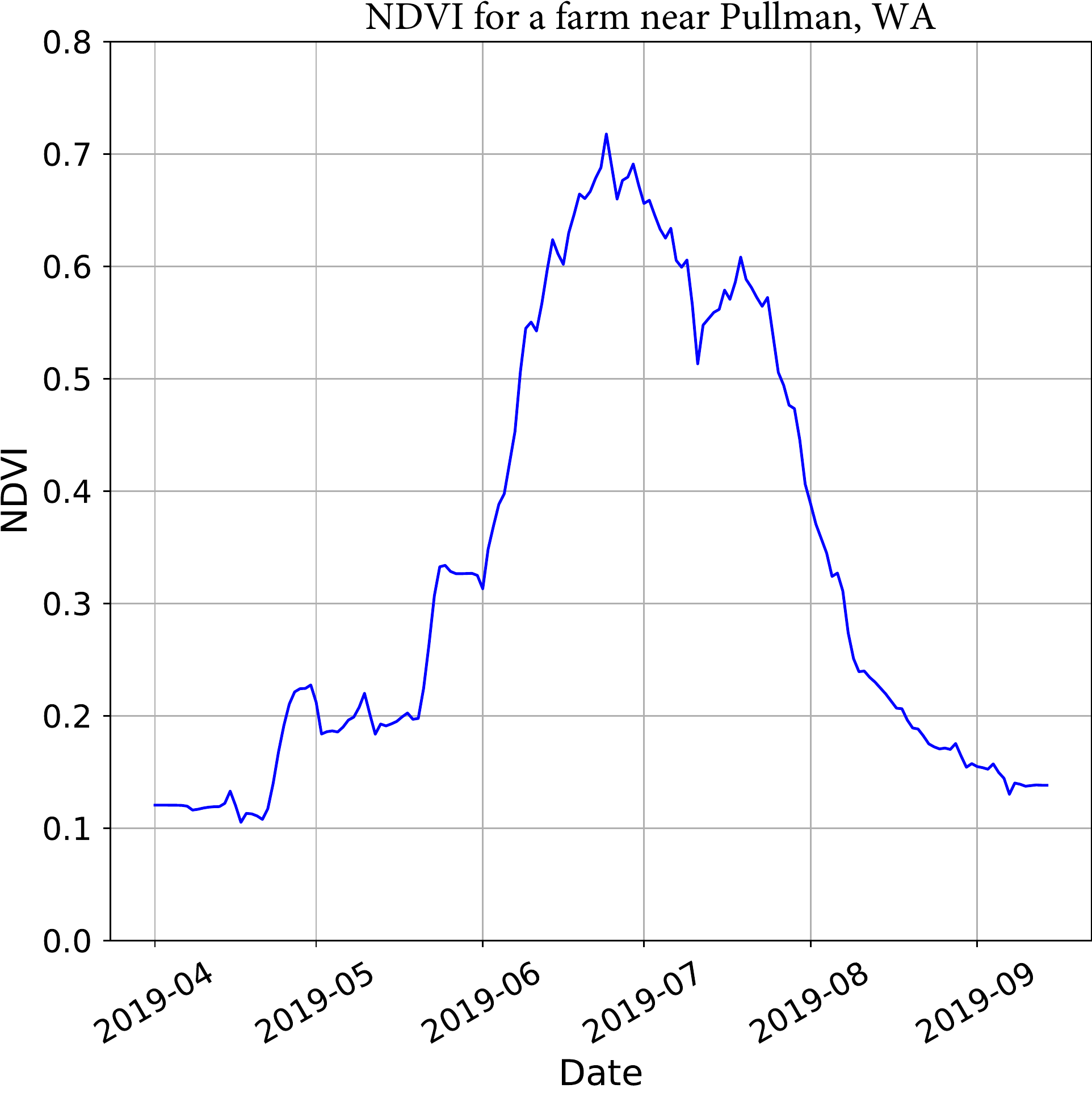} \hspace{1cm} \includegraphics[width=0.4\linewidth]{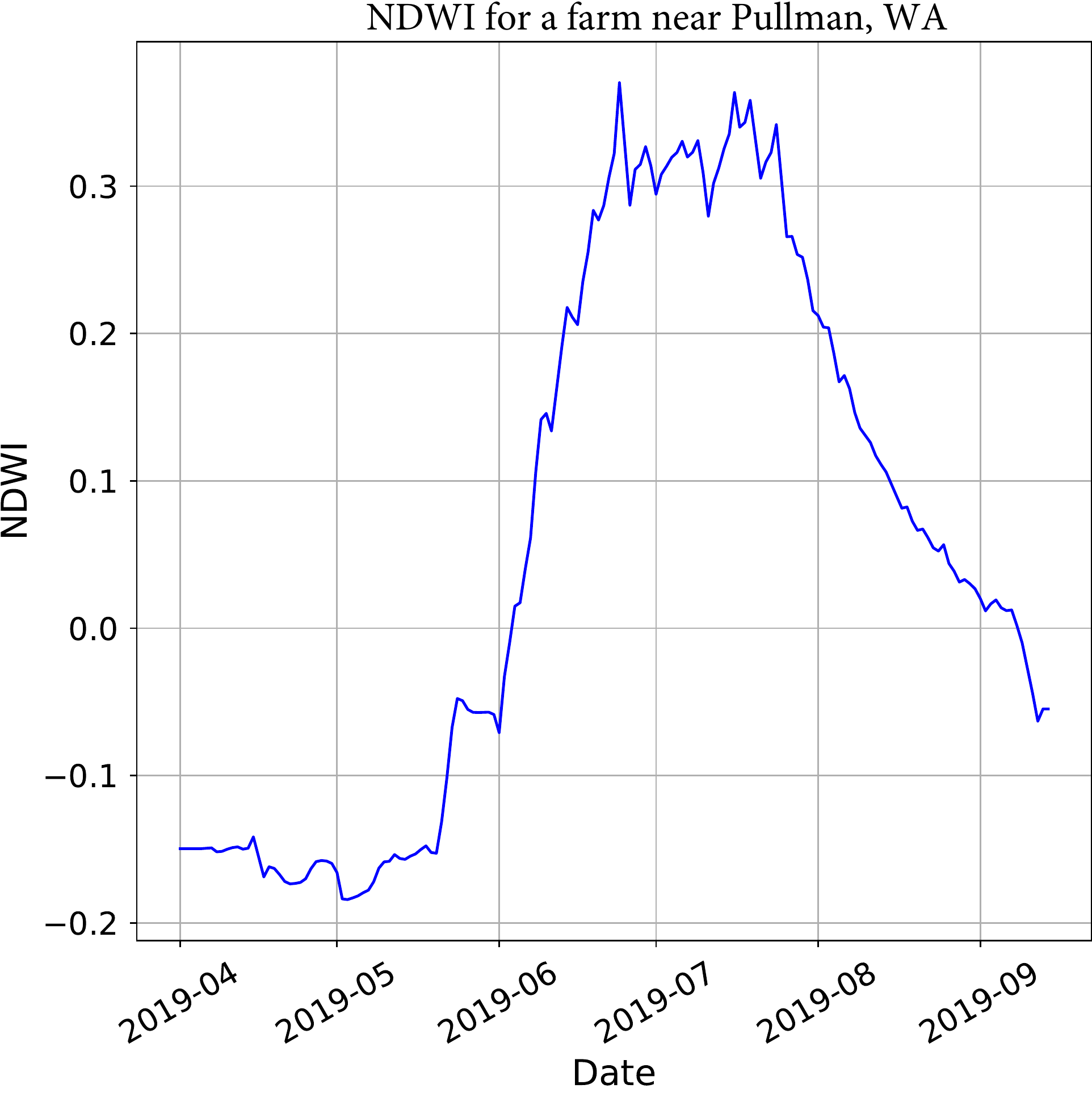}
\caption{
NDVI and NDWI predictions values as a function of time for a hilly farm for the 2019 growing season near Pullman, Washington.  
}
\label{fig:butte_ndi_plot}
\end{figure}

In this section, we give further pictorial insight into the algorithms discussed in the paper.  
First, we apply \name~ to predict the normalized difference vegetation index (NDVI) and the normalized difference water index (NDWI) first for a single farm plot and then for a large area, both in Washington state.  The farm is situated near Pullman, WA, and sits on top of a butte.  We can see the topology of the field reflected in variations of the NDVI and NDWI.  In Figure~\ref{fig:butte_s2} we see the available Sentinel-2 imagery for a period from 4/1/2019 until 9/15/2019.  A total of 6 Sentinel-2 images are available for each 15 days, but many are covered in clouds, and we have masked out pixels not belonging to the farm.    Figure~\ref{fig:butte_fake} shows daily predictions using \name~ for the same time period, while Figure~\ref{fig:butte_ndvi} and Figure~\ref{fig:butte_ndwi} shows the derived daily NDVI and NDWI values from \name.  We see in these images how the vegetation matures through the season, then dries off and is eventually harvested.  
We also see variations across the farm, both for the NDVI and NDWI.  Figure~\ref{fig:butte_ndi_plot} shows the average NDVI and NDWI as a function of time as predicted by \name.  NDWI values below $0$ is a good indicator that there is very little water content in the crop.  Consequently, the NDWI images in Figure~\ref{fig:butte_ndwi} could have been used as a guide to decide on when to harvest.

\begin{figure}
\centering
  \includegraphics[width=0.4\linewidth]{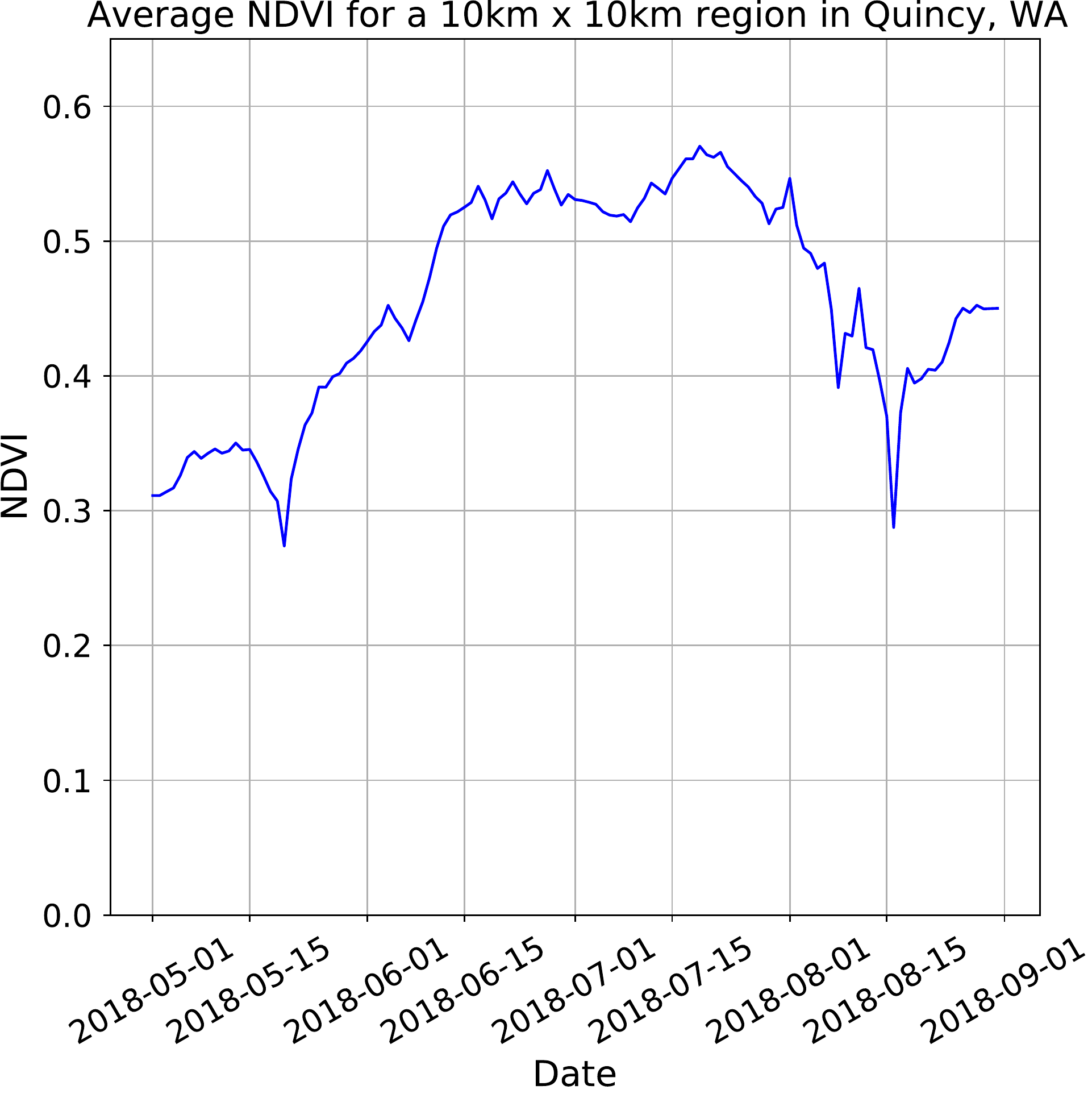}\hspace{1cm}\includegraphics[width=0.4\linewidth]{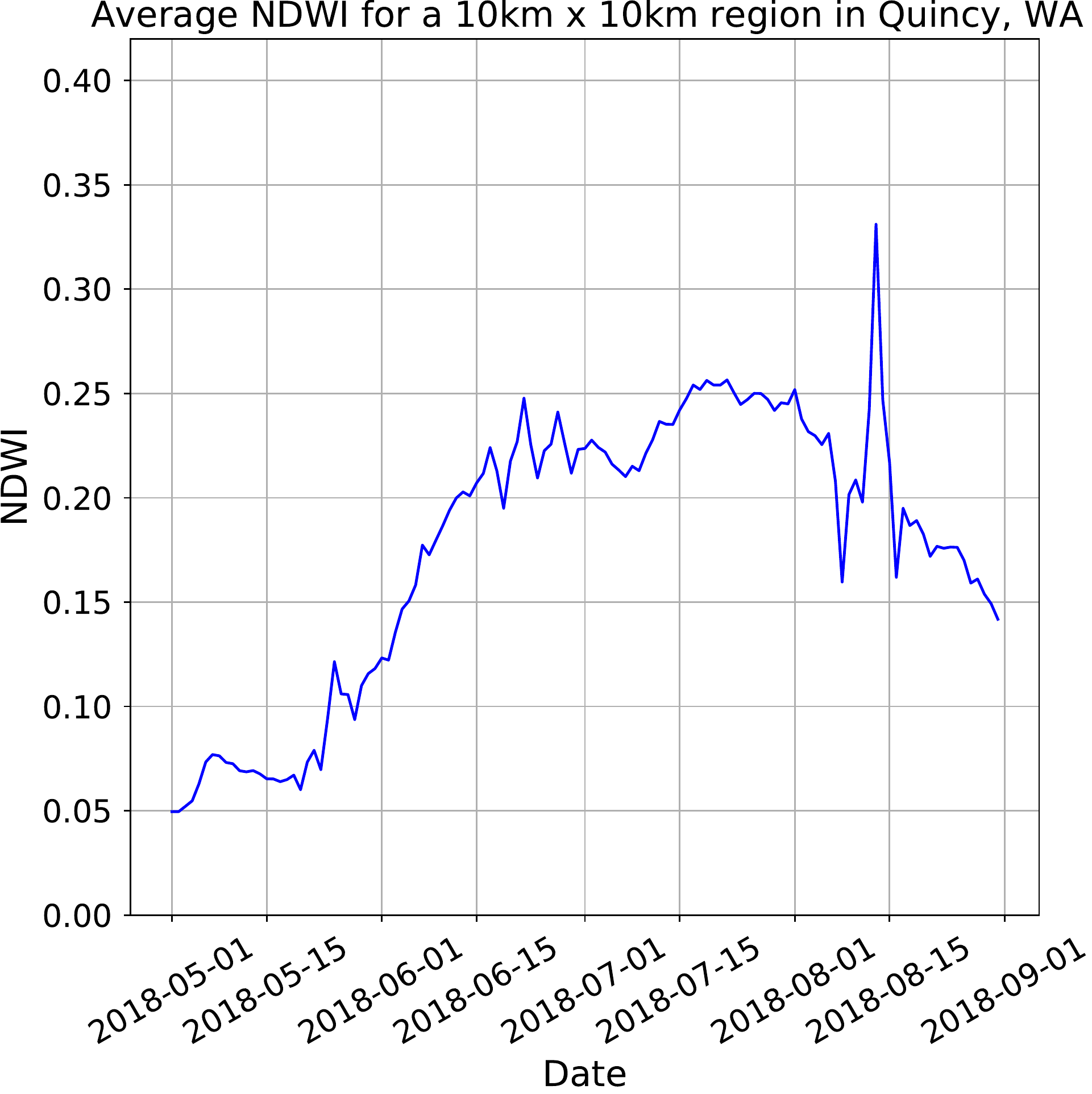}
\caption{
Average NDVI and NDWI values for a large 10km $\times$ 10km region in Quincy, Washington.  
}
\label{fig:quincy_ndi_plot}
\end{figure}

\begin{figure*}
\centering
  \includegraphics[width=\linewidth]{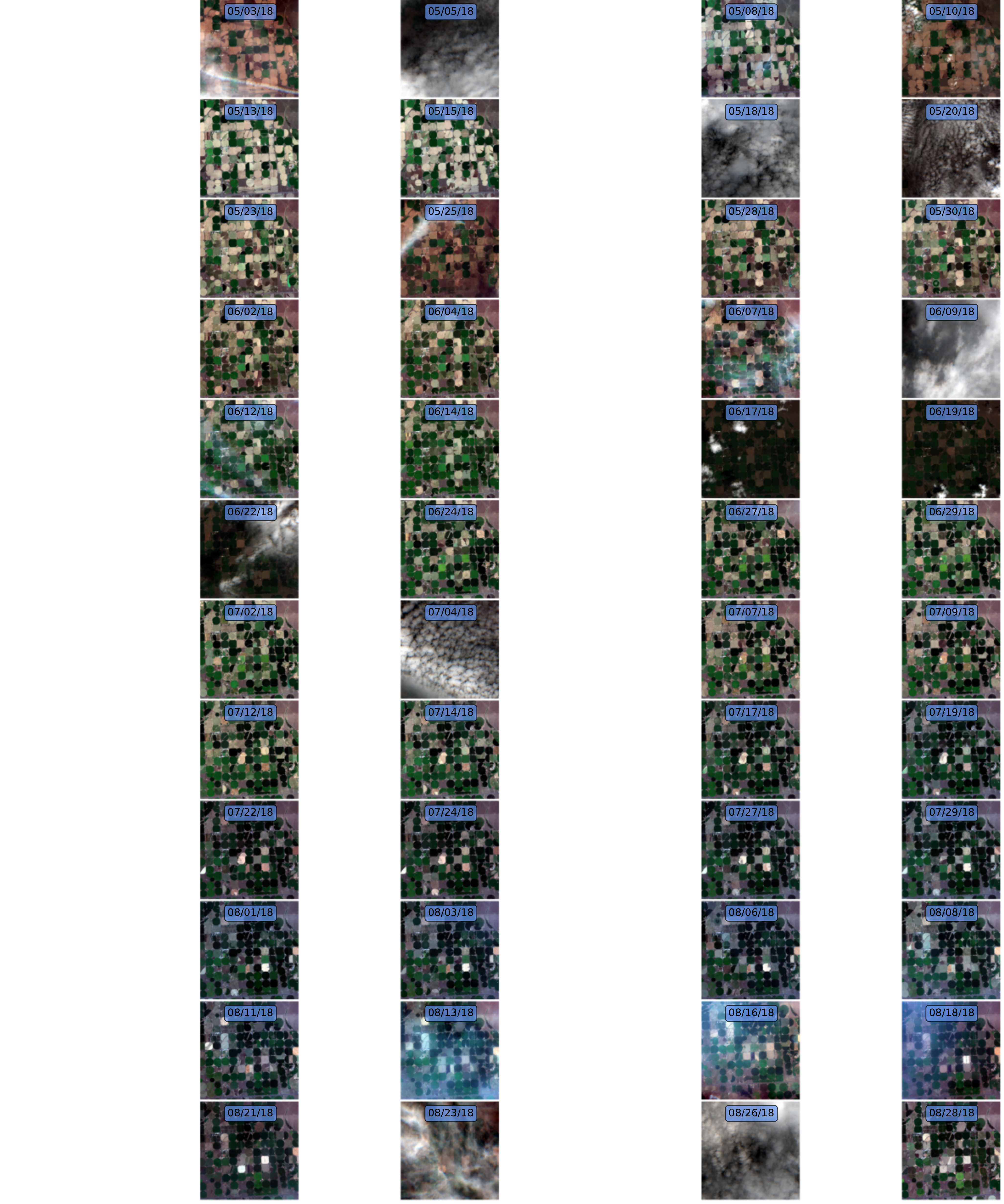}
\caption{
120 days of Sentinel-2 images for a 10km $\times$ 10km region in Quincy, Washington.
}
\label{fig:quincy_s2}
\vspace{-10pt}
\end{figure*}

\begin{figure*}
\centering
  \includegraphics[width=\linewidth]{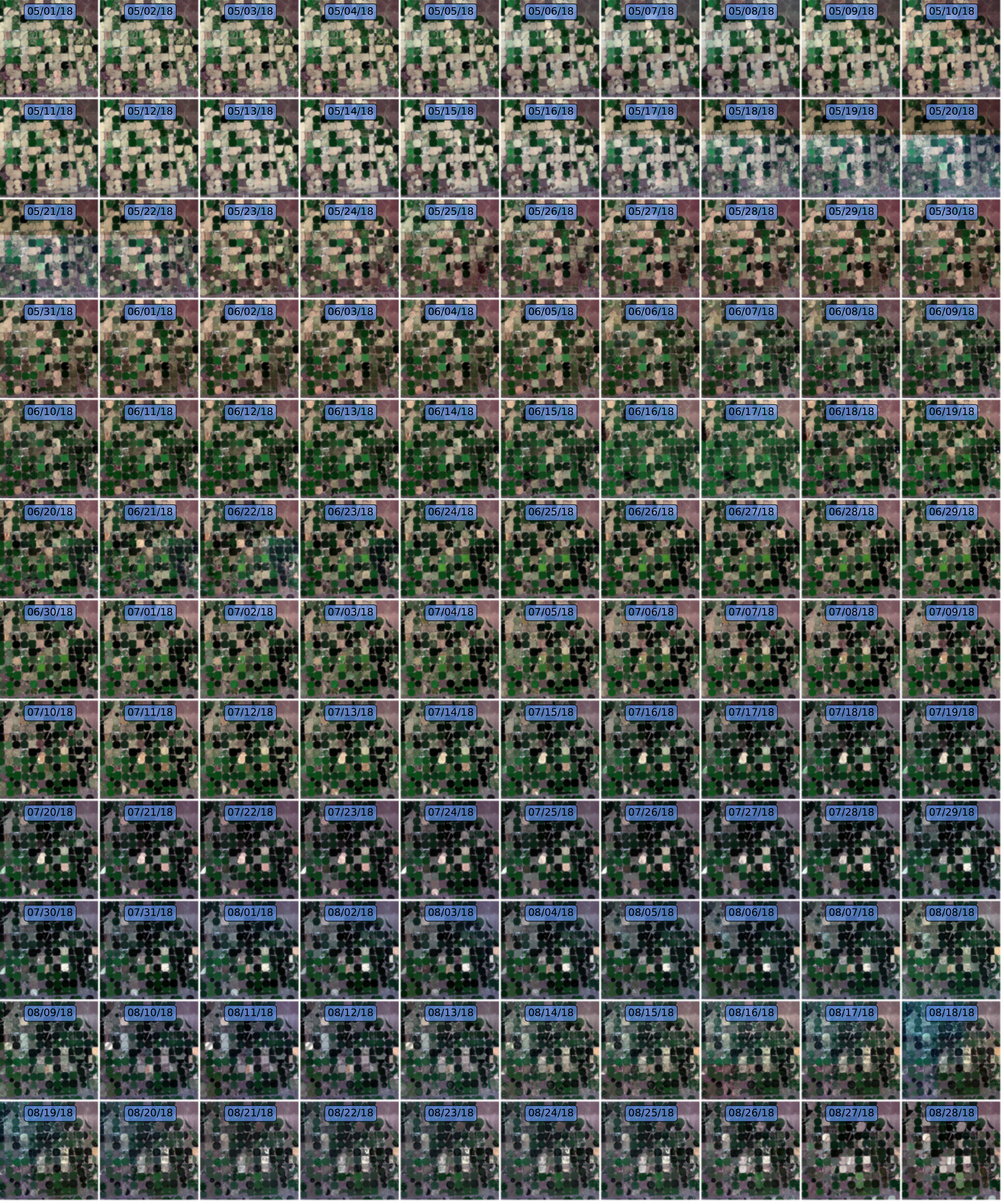}
\caption{
120 days of VIR predictions with \name~ for a 10km $\times$ 10km region in Quincy, Washington.
}
\label{fig:quincy_rgb}
\vspace{-10pt}
\end{figure*}

\begin{figure*}
\centering
  \includegraphics[width=\linewidth]{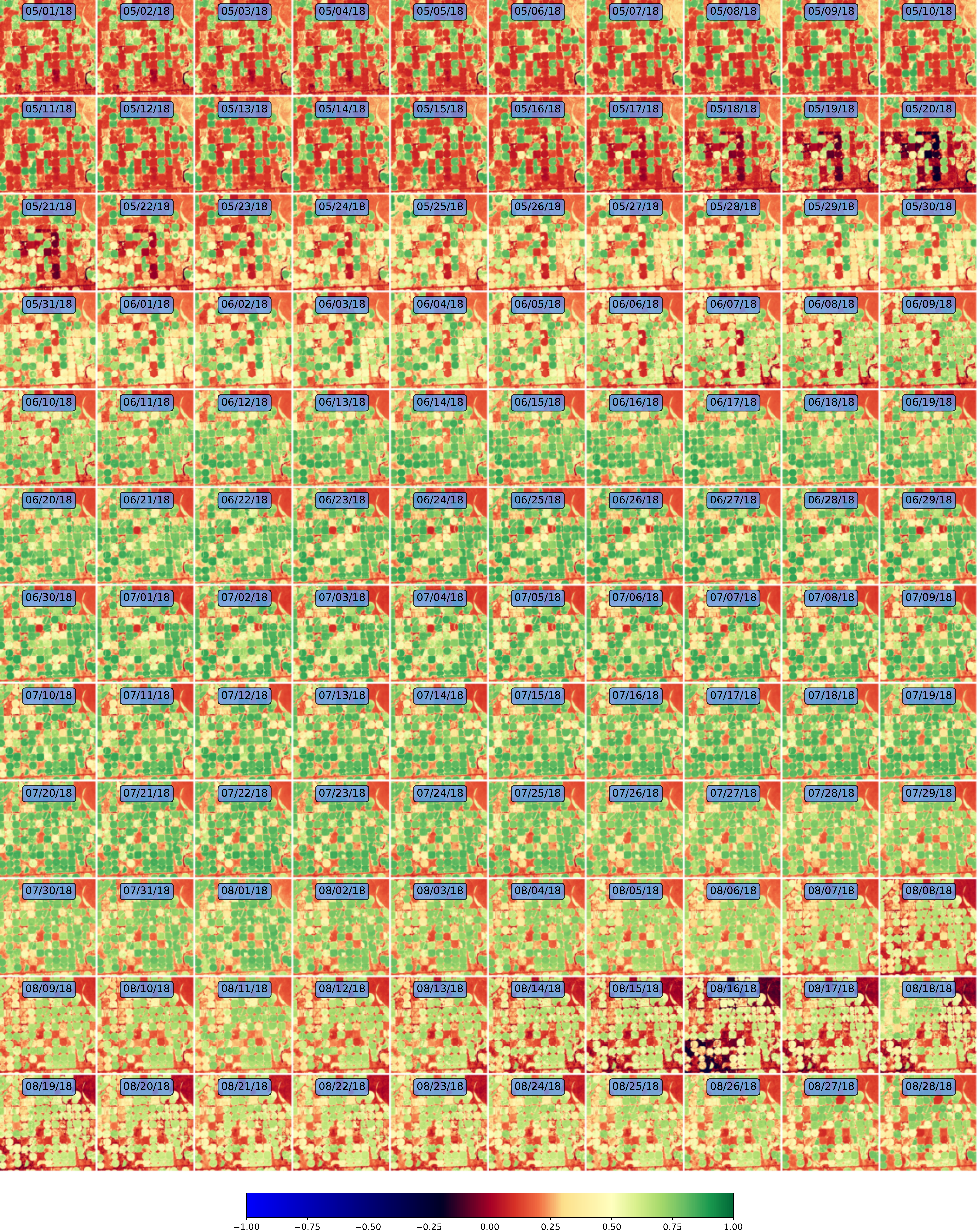}
\caption{
120 days of NDVI predictions with \name~ for a 10km $\times$ 10km region in Quincy, Washington.
}
\label{fig:quincy_ndvi}
\vspace{-10pt}
\end{figure*}

\begin{figure*}
\centering
  \includegraphics[width=\linewidth]{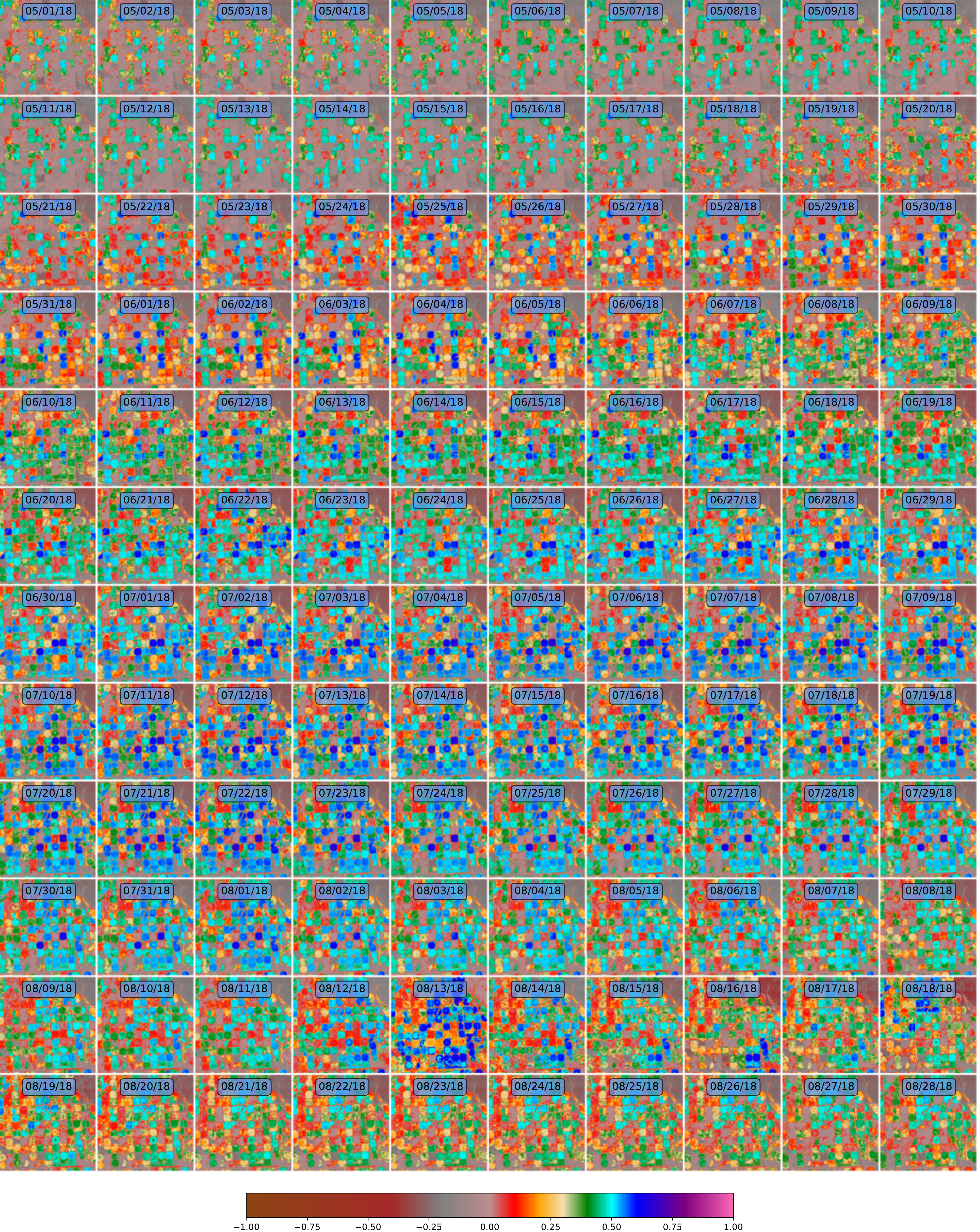}
\caption{
120 days of NDWI predictions with \name~ for a 10km $\times$ 10km region in Quincy, Washington.
}
\label{fig:quincy_ndwi}
\vspace{-10pt}
\end{figure*}

The same series of plots were also repeated for a larger 10km $\times$ 10km region in Quincy, WA.  This is a relatively dry area, where much of the crop is irrigated, as seen from the circular fields with pivoting irrigation.  Since some of the scenes are not cultivated, and the average NDVI and NDWI values are lower, making for a much smaller peak in Figure~\ref{fig:quincy_ndi_plot}. Figures~\ref{fig:quincy_s2}, \ref{fig:quincy_rgb}, \ref{fig:quincy_ndvi} and \ref{fig:quincy_ndwi} shows respectively the available Sentinel-2 images, daily predictions by \name, NDVI predictions by \name~ and NDWI predictions by \name.  We can see when the fields get harvested as the NDVI becomes red and the NDWI gray.


\section{Satellite Data Processing}\label{sec:data}
In this section, we review the data processing steps used with \name~ for both the SAR and the multispectral data.  We used ESA's Sentinel-1 \citeA{torres2012gmes}  for SAR and Sentinel-2 \citeA{drusch2012sentinel} for multispectral, both of which provide open access as governed by the "Legal Notice on the use of Copernicus Data and Service Information."

\subsection{Multispectral Data Processing}\label{subsec:Sentinel2}
We used multispectral data from the Sentinel-2 satellite system.  Sentinel-2 is a constellation of the twin satellites, Sentinel-2A and Sentinel-2B, launched into orbit in June 2015 and March 2017, respectively.  The satellites are in a 10 day orbital cycle with a joint repeat cycle of 5 days.  Each pass images a swath width of 290km, and the temporal coverage increases for locations inside overlapping swaths.  These satellites have a joint revisit time of 5 days and systematically images land surfaces between $56^{\circ}$S and $84^{\circ}$N.  The Sentinel-2 products are provided in tiles aligned with the military MGRS zones.  Each tile or granule covers roughly 100x100km of land and is projected onto one of 60 Universal Transverse Mercator (UTM) coordinate systems.

The multispectral data available for Sentinel-2 contains 13 spectral bands with resolutions of $10m\times 10m$ for the three visible (red-green-blue) and near infra-red (NIR) bands, $20m\times 20m$ for the three vegetation red edge (VRE), narrow near-infrared (NNIR), and the two short-wave infra-red (SWIR) bands and $60m\times 60m$ for the aerosol, water vapour, and cirrus bands.  The resolution mismatch, lack of cloud shadow mask and other artifacts, such as image artifacts on edge and a cloud mask not ideal for \name~ led us to process the original data available from Sentinel-2 into an analysis-ready format for training and inference.

\header{Data Access} We used the python \texttt{sentinelsat} module \citeA{sentinelsat} for programmatic access to the Copernicus Open Access Hub \citeA{Copernicus}.  This tool was used for data not older than 1 year at the time of access.  It can also be used for historical data, but only one product can be requested every 30 minutes.  Alternatively, historical data can also be purchased from approved Copernicus vendors \citeA{sentineldataaccess}.

\header{Image Fusion}
The Sentinel-2 products provided do not always cover an entire tile (aka granule), and sometimes several images will cover different sections of the granule.   When these images were temporally close (seconds apart) and on the same absolute orbit, we combined the images using \citeA{gdal_merge}.  Generally speaking, we see no apparent artifacts in the merged image for cloud-free pixels, but the clouds may have shifted between the merged images.

Additionally, we also merged all images in a window for the time-step requested by the neural network (1 or 2 days).  For a time step of 1 day, there were no additional image merges, while for time steps larger than 1 day we selected the first image with the most cloud-free pixels.  The remaining images were either discarded or shifted to an image-free neighboring time-step.  For regions with extremely high cloud densities, where there are no cloud-free pixels in a 48 day period, we utilized the time step of 2 days to process 96 days at a time.\footnote{For the test set we reported results for, only a time step of 1 day was utilized.}

\header{Stacking and Resampling}  
Sentinel-2 products contain 13 spectral bands.  Four bands (red, green, blue, and near-infrared (NIR)) are provided at a $10m\times 10m$ pixel resolution, six bands (three VRE bands, NNIR and two SWIR) are provided at $20m\times 20m$ pixel resolution, while the last three bands (Aerosol, Water Vapor, and Cirrus) are at a $60m \times 60m$ pixel resolution.  In order for the training algorithms to be efficient, we resampled all bands using bicubic interpolation to a common $10m \times 10m$ pixel resolutions and stacked them in a single \texttt{GEOTIFF} file.  The \texttt{GEOTIFF} used the \texttt{TILED=YES} option with the default block-size of 256 to allow for efficient retrieval of rectangular sub-sections of the full image.  This format is similar to the Cloud Optimized \texttt{GEOTIFF} (COG) standard \citeA{durbin2020task}, but lacking the overview capability.  We plan to comply with the COG standard in the future.

\header{Missing Data} In tiles that intersect the edge of an individual data strip, there will be missing pixels.  We treat missing pixels as if they were clouds.  Specifically, we created a "cloud-mask" that combined cloud shadows, clouds, and missing pixels into a single mask.  

\header{Edge Effects}  
Due to the particular geometrical layout of the focal plane, each spectral band of the multispectral instrument observes the ground surface at different times, \citeA{sentinel2msi}.  In particular, the extent captured for each spectral band differs, yielding an undesirable rainbow-like effect on the edge of the images.  We removed these regions by considering all pixels with a \texttt{nodata value} in {\em any} band as a missing pixel to be masked out.

\subsection{Detecting clouds in satellite images}\label{sec:clouddetection}
 Clouds interfere with atmospheric correction algorithms, frequently occlude the surface, and add noise to surface observations. Thus, one of the first steps in processing satellite images is a cloud segmentation operation.  Cloud shadows are another source of noise, and cloud shadow segmentation can additionally utilize the cloud locations, the solar illumination angle, and the satellite viewing angle.

Cloud detection and cloud-shadow detection are well-studied problems, \citeA{foga2017cloud, sanchez2020comparison, bai2016cloud, jeppesen2019cloud}.  Cloud detection can be divided into pixel-wise, single-scene, and multi-temporal algorithms.  Pixel-wise algorithms are threshold-based and use the entire available spectrum as in the case of Function of Mask (Fmask) \citeA{zhu2012object}, now in its fourth version \citeA{qiu2019fmask}.  Fmask is widely used and additionally provides a cloud-shadow mask as well.  The cloud detection model used in this paper was based on the \texttt{S2Cloudless} algorithm, \citeA{baetens2019validation}. \texttt{S2Cloudless} is another pixel-wise thresholding that uses the LightGBM decision tree model \citeA{ke2017lightgbm}.  
 
Single-scene algorithms can utilize the spatial context through deep learning approaches.  Deep cloud detection largely relies on the U-net architecture and has been shown to outperform the Fmask algorithm, \citeA{jeppesen2019cloud, zhang2019cloud, guo2020cloud}.  The deep learning approach requires a large annotated data set.  Such a data set is, however, not publicly available for Sentinel-2.  \citeA{mohajerani2019cloudmaskgan} proposed using a GAN to artificially generate realistic cloudy satellite images that could possibly be used to train such a model. 
 
 The most accurate algorithms are in the multi-temporal context.  For satellites such as Landsat, Sentinel, and MODIS with predictable revisit rates, this is the most accurate algorithm class.  The algorithms are computationally more complex and costly but offer the highest performance.  The state-of-the-art performance for Sentinel-2 is the MAJA algorithm \citeA{lonjou2016maccs}.  Multi-temporal cloud detection has also been done for the Pl{\'e}iades satellite \citeA{champion2012automatic}, where a less extensive temporal context is available.  

\header{Cloud Annotation}
Sentinel-2 is a relatively new product compared to MODIS and Landsat, so there are few open datasets available for cloud and cloud shadow annotations.  One available source was the Hollstein dataset \citeA{hollstein2016ready}.  It contained a list of pixels with one of 6 classes annotated (cloud, cloud-shadow, clear, cirrus, snow, and water).  Without the spatial and temporal context available, this meant that we were constrained to the pixel-wise algorithms by using the Hollstein dataset.

\header{Cloud Detection} 
The Hollstein data-set was used in building a pixel-wise LightGBM \citeA{ke2017lightgbm} decision-tree based model called \texttt{S2Cloudless}, \citeA{baetens2019validation} for detecting clouds.  We used the very efficient and light-weight \texttt{S2Cloudless} algorithm as a basis for our cloud-detector.  The main problem with using the algorithm as-is, was that persistently white object-like buildings was masked out in images for all time points, which guaranteed failure of the reconstruction algorithm at these locations.  As noted in \citeA{baetens2019validation}, the most accurate cloud detection algorithms like the state-of-the-art \citeA{MAJA} requires temporal sequences. In our augmented model, we simply removed connected regions of clouds and cloud-free regions when these regions had less than 400 pixels, and we also removed pixels that \texttt{S2Cloudless} marked as cloudy much more frequently than the Sentinel-2 native cloud-mask.  We generated a cloud-mask for every Sentinel-2 tile prior to training and inference with \name.  This leads to a pleasant user experience with very little lag time when running \name~ on small space-time regions.

\header{Cloud Shadow Detection} 
We built our own pixel-wise cloud-shadow detector using the ExtraTrees classifier \citeA{geurts2006extremely} provided in \texttt{Scikit-learn} and the Hollstein dataset.  The resulting random forest had a high cloud-shadow detection recall (99\%) but a low precision.  Consequently, we selected only the pixels that were significantly darker (90\%) than the average cloud-free luminance for the mean of the same pixel in a window of the 20 temporarily closest Sentinel-2 images.   The resulting algorithm fails to detect topological shadows or cloud shadows for periods with a prolonged cloud cover.
Our cloud and cloud-shadow detectors are far from state-of-the-art and a potential target for future work. 

\subsection{SAR Data Processing}\label{subsec:sar-processing}
We used radar data from the Sentinel-1 satellite system.  Sentinel-1 is a constellation of two satellites, Sentinel-1A and Sentinel-1B,  launched into low-earth-orbit in April 2014 and 2016, respectively.  The satellites are in a 12 day cycle with a $180^{\circ}$ orbital phasing difference, giving a 6 day repeat frequency.  The location revisit rate is 3 days at the equator and 2 days in Europe.  For our experiments, we used data from the interferometric wide (IW) swath mode that covers 250km with dual polarisation VV-VH.  The original data needs substantial processing to be transformed into an image that can be used in our system.  The original pixel resolution is $20m\times 5m$ in the azimuth and range directions.  

A significant number of steps were needed in order to process data from the Sentinel-1 into images that can be used by \name.  The rest of this section outlines the steps. 

\header{Data Access} We used the python \texttt{sentinelsat} module \citeA{sentinelsat}, to download recent ground range detected (GRD), dual polarisation Level-1  Sentinel-1 products.  For data older than 1 year, data was acquired through the services of Alaska Satellite Facilities \citeA{asf}.    

\header{SAR processing}  We used the \texttt{SNAP} toolkit \citeA{snap} to generate image data from the downloaded data.  The processing steps were as follows:
\begin{description}
\item[$\texttt{do\_apply\_orbit\_file}$] This step acquires the satellite orbit file, which is needed later.  The precise orbit file is only available 20 days afterward, but processing with a less precise orbit is still possible.  We reprocess all images after the more precise orbit file becomes available.
\item[$\texttt{do\_remove\_GRD\_border\_noise}$] The Sentinel-1 GRD product has noise artifacts at the image borders, which are quite consistent at both the left and right sides of the satellite’s cross-track and at the start and end of the data take along the track.  The software we used is described and compared to alternatives in \citeA{ali2018methods}.
\item[$\texttt{do\_thermal\_noise\_removal}$] Sentinel-1 image intensity is disturbed by additive thermal noise.  Thermal noise removal reduces noise effects in the inter-sub-swath
texture \citeA{filipponi2019sentinel}.
\item[$\texttt{do\_calibration}$] Calibration converts pixel values to radiometrically calibrated SAR back-scatter.
\item[$\texttt{do\_terrain\_correction}$] Range Doppler terrain correction rectifies geometric distortions caused by topography, such as foreshortening and shadows.  It uses a digital elevation model to correct the location of each pixel.  The methods used implements the Range Doppler orthorectification method \citeA{small2008guide}.  This step also resamples the resolution to $10m\times 10m$ for alignment with corresponding Sentinel-2 products. 
The default setting for the terrain correction is to remove ocean pixels.  We explicitly disabled this to use the same algorithm for both land and sea locations. 
\item[$\texttt{linear\_to\_db}$]  This simply brings the signal to the log-domain.
\item[$\texttt{do\_subset}$] In this step, we extract the extent of the image that falls inside each intersecting Sentinel-2 tile and apply the corresponding UTM projection.
\end{description}

\header{Image Fusion}  
The Sentinel-1 product contains 2 bands corresponding to VH and VV polarisations.  We stacked the VV and VH bands and compressed the tiff files using the loss-less Z standard (ZSTD) compression algorithm.  We also use the \texttt{TILED=YES} option as before.  Additionally, we merged images along the same orbital sweep and marked pixels outside the image as missing.

\subsubsection{Dataset Characteristics}
Sentinel-1 SAR utilizes C-band radar which partially penetrates canopy, dry sand, and thin layers of snow, while it is reflected off of standing water.  As a result, the resulting SAR images give a lot of useful information for agriculture and are ideally suited to detect flooding.  The same reflective property also means that changes in watercolor cannot be detected solely from SAR and has to also rely on multispectral data.  As a result, our system is not well suited to monitor water quality. Similarly, snowy ground can be challenging for the system (however, our baselines perform even worse on snowy scenes).

\section{Damped Interpolation and Matrix Completion}\label{subsec:tc}
In this section, we outline the optimization methodology used for the damped interpolation and the matrix completion objective used in the paper.  One of our requirements for the methods is that they should run efficiently so that the methods could be used interactively.  This was achieved mainly by executing the methods on the GPU using \texttt{pytorch}.

\subsection{Optimization for Damped Interpolation}
Recall that the objective function is given by
\begin{equation}\label{eqn:mc}
   F(\mX) = \|\mM \circ (\mX-\mY)\|_F^2 + \alpha \sum_{t=1}^{T-1} \|\mX_{t+1}-\mX_t\|_F^2,
\end{equation}
where $\mY\in\R^{(C_1+C_2) T\times HW}$ is the stacked multispectral and radar data, $\mM\in\R^{(C_1+C_2) T\times HW}$ is the data mask (0 for missing data and cloudy pixels, 1 for clear sky pixels), and $\mX\in\R^{(C_1+C_2) T\times HW}$ is the desired cloud-free reconstruction.  The notation $\mX_t$ is used to refer to the smaller sub-matrices of size $(C_1+C_2) \times HW$ corresponding to the data at time $t$.

With even the modest choice of dimensions $C_1=10$, $C_2=2$, $T=48$, $H=W=448$ $F$ is a quadratic objective in $1.2*10^8$ variables which is already too large for a general quadratic solver.  We consider a formulation that takes advantage of the problem structure and allows an iterative approach with much less expensive steps.  The unconstrained version of this problem decouples into one problem for each pixel, band, and each can be solved separately.  However, we consider a solution where larger matrix operations can be done, as each of the pixel band problems would have its own structure.

\header{The Damping Coefficient:}
The function $F$ has the interesting property that $\lim_{\alpha\rightarrow 0^+} \mathrm{argmin}_{\mX}F(\mX)$ converges to a function that linearly interpolates the temporal sequence of $\mY$ with a constant function as the extrapolation before the first cloud-free pixel and after the last cloud-free pixel.  In other words, it is equivalent to a solution that does pixel-based linear interpolation.  For $\alpha>0$ it pulls the cloud-free observations closer towards each other, thus the name of the method.  We refer to $\alpha$ as the damping coefficient.  

\header{Auxilliary function:}  We used an auxilliary function that gives an upper bound to the objective function $F$.
\begin{eqnarray*}
Q(\mZ,\mX) &=& \|\mM\circ(\mX-\mY)\|_F^2 + \|(1-\mM)\circ(\mX-\mZ)\|_F^2 \\
& & + \alpha\sum_{t=1}^{T-1} \|\mX_{t+1}-\mX_t\|_F^2\\
&=& \|\mX- \mM\circ \mY-(1-\mM)\circ (\mZ)\|_F^2 \\
& & + \alpha\sum_{t=1}^{T-1} \|\mX_{t+1}-\mX_t\|_F^2.\\
\end{eqnarray*}
We have $Q(\mX,\mX) = F(\mX) $ and $Q(\mX,\mZ)\geq F(\mX)$ and consequentially the global minimum of $Q$ coincides with that of $F$.  We use an alternate minimization over $\mX$ and $\mZ$ to iteratively reduce the value of $Q(\mX,\mZ)$.

For fixed $\mX$ the minimum over 
$\mZ$ is achieved for  $\mZ=(1-\mM)\circ\mX$.  
The minimization over $\mX$ for fixed $\mZ$ turns out to also be simpler than before.  The independent pixel problems can be solved in parallel.  

\header{The minimizer:}
To describe the solution define the forward difference matrix $\mD\in\R^{T\times T}$ by
\begin{equation}
\mD = \begin{pmatrix} 
-1 & 1 & 0 & \cdots & 0 & 0 \\
0 & -1 & 1 & \cdots & 0 & 0 \\
0 & 0 & -1 & \cdots & 0 & 0 \\
\vdots & \vdots & \vdots & \ddots & \vdots & \vdots \\
0 & 0 & 0 & \cdots & -1 & 1 \\
0 & 0 & 0 & \cdots & 0 & 0 \\
\end{pmatrix}.
\end{equation}
and the companion matrix $\mDelta=\mD\otimes \mI_{C_1+C_2}\in\R^{T(C_1+C_2)\times T(C_1+C_2)}$.
Fixing $\mZ$ the minimum for $\mX$ is then given by
$$
\mX=(\mI+\alpha\mDelta^\top\mDelta)^{-1} ((\mM\circ \mY)-(1-\mM)\circ\mZ).
$$
Furthermore, due to the properties of the Kronecker product we have
$$
(\mI+\alpha\mDelta^\top\mDelta)^{-1} = (\mI_{N_t}+\alpha \mD^\top \mD)^{-1} \otimes \mI_{N_b}.
$$
Since $\mI_{N_t}+\alpha \mD^\top \mD$ only depends on the dimension $T$ and $\alpha$ it can be pre-computed, so that all calculations are simple matrix multiplies that can easily be handled on the GPU. Putting the resulting two iterations together we get a single iteration:
\begin{equation}
\mX\longleftarrow (\mI+\alpha\mDelta^\top\mDelta)^{-1}  ((\mM\circ \mY)-(1-\mM)\circ\mX).
\end{equation}

\subsection{Optimization for Matrix Completion}
Several approaches for cloud removal has used matrix and tensor completion with non-smooth optimization,  \citeA{ji2018nonlocal,aravkin2014variational, ma2017fusion}.  We aimed for simpler, more efficient models and based our work on the methods in \citeA{wang2016removing}.  Their approach used the nuclear norm, which although a convex function, is quite complex and slow to optimize.  We simplified it to a low-rank quadratic optimization problem which was solved with an alternating optimization algorithm implemented on the GPU.  Both the damped interpolation and the matrix completion methods are based on the same objective loss function, but the matrix completion method enforces a rank constraint.  \citeA{wang2016removing} considered the bands to be independent, but combining bands allows the solution to have a higher rank and benefits from the use of the SAR data.  

$\mX$ can be decomposed into a low-rank matrix $\mX=\mU\mV^\top$ with $\mU\in\R^{(C_1+C_2)T\times N_r} , \mV \in\R^{HW\times N_r}$, where $N_r$ is the rank.  
We think of the column vector $V_r$ as land-type $r$ and the column vector $U_{r}$ as the evolution of the corresponding land-type.  Our matrix completion minimizes $F(\mU\mV^\top)$ with respect to $\mU$ and $\mV$ for a fixed rank $N_r$.  We found the best results with $N_r=35$, which is considerably smaller than the full rank $(C_1+C_2)T=576$ for our test set ($T=48$).  The optimization algorithm was slower than the damped interpolation and relied on alternating optimization as well.

\header{The minimizer:}
We use the same auxiliary function as in the previous section and define $\mY_Z=((\mM\circ \mY)-(1-\mM)\circ\mZ)$.  We then optimize $G(\mU,\mV,\mZ)=Q(\mU\mV^\top,\mZ)$ iteratively with respect to $\mZ$, $\mV$ and $\mU$.  Since all the equations are quadratic with the other variables fixed we get closed form equations:
\begin{eqnarray}
\mZ &=& (1-\mM) \circ (\mU\mV^\top)\\
\mU &=& (\mI+\alpha\mDelta^\top\mDelta)^{-1} \mY_Z\mV (\mV^\top\mV)^{-1}\\
\mV &=& \mY_Z^\top\mU(\mU\mU^\top+\alpha\mU^\top\mDelta^\top\mDelta\mU)^{-1}.
\end{eqnarray}
Note that with the exception of $(\mI+\alpha\mDelta^\top\mDelta)^{-1}$ which can be pre-computed, the matrix inverses are of size $N_r\times N_r$ and we explicitly control the size of $N_r$.  The corresponding loss with the nuclear norm regularizer 
$F(\mX)+\lambda\|\mX\|_*$ can be done in the same manner by using the auxilliary identity $\|\mX\|_*=\min_{\mU,\mV: \mX=\mU\mV^\top}\frac{1}{2}(\|\mU\|_F^2+\|\mV\|_F^2)$.  However, after the dust settles the auxilliary formulation for the nuclear norm still requires solving a Sylvester equation that is not available in \texttt{pytorch} for the GPU.

\bibliographystyleA{abbrv}
\bibliographyA{references}

\end{document}